\documentclass{article}

\usepackage[preprint]{neurips_2025}

\usepackage[utf8]{inputenc} % allow utf-8 input
\usepackage[T1]{fontenc}    % use 8-bit T1 fonts
\usepackage{hyperref}       % hyperlinks
\usepackage{url}            % simple URL typesetting
\usepackage{booktabs}       % professional-quality tables
\usepackage{amsfonts}       % blackboard math symbols
\usepackage{nicefrac}       % compact symbols for 1/2, etc.
\usepackage{microtype}      % microtypography
\usepackage{xcolor} 
\usepackage{algorithm}% colors
\usepackage{algorithmic}
\usepackage{amsmath}
\usepackage{graphicx}
\usepackage{multirow} 
\usepackage{subcaption}
\usepackage{wrapfig}
\usepackage{setspace}
\usepackage[capitalize]{cleveref}

\crefname{section}{Sec.}{Secs.}
\Crefname{section}{Section}{Sections}
\Crefname{table}{Table}{Tables}
\crefname{table}{Tab.}{Tabs.}
\crefname{algocf}{alg.}{algs.}

\title{Adversarially Robust AI-Generated Image Detection for Free: An Information Theoretic Perspective}

\author{%
  \textbf{Ruixuan Zhang}$^{1}$,
  \textbf{He Wang}$^{2}$,
  \textbf{Zhengyu Zhao}$^{3}$,
  \textbf{Zhiqing Guo}$^{4}$,\\
  \textbf{Xun Yang}$^{5}$,
  \textbf{Yunfeng Diao}$^{1}$,
  \textbf{Meng Wang}$^{1}$ \\
  $^1$Hefei University of Technology \quad
  $^2$University College London \quad
  $^3$Xi'an Jiaotong University \\
  $^4$Xinjiang University \quad
  $^5$University of Science and Technology of China \\
}

\begin{document}

\maketitle

\begin{abstract}
Rapid advances in Artificial Intelligence Generated Images (AIGI) have facilitated malicious use, such as forgery and misinformation. 
Therefore, numerous methods have been proposed to detect fake images.
Although such detectors have been proven to be universally vulnerable to adversarial attacks, defenses in this field are scarce.
In this paper, we first identify that adversarial training (AT), widely regarded as the most effective defense, suffers from performance collapse in AIGI detection.
% More importantly, the underlying cause remains unclear.
% we theoretically and empirically demonstrate that adversarially trained detectors suffer from feature entanglement, 
Through an information-theoretic lens, we further attribute the cause of collapse to feature entanglement, which disrupts the preservation of feature–label mutual information.
% thereby degrading discriminative performance.
% Unlike the feature entanglement caused by AT, standard trained detectors show that adversarial features are clearly separated from clean ones with significant feature shifts.
Instead, standard detectors show clear feature separation.
Motivated by this difference, we propose \textbf{T}raining-free \textbf{R}obust Detection via \textbf{I}nformation-theoretic \textbf{M}easures (\textbf{TRIM}), the first training-free adversarial defense for AIGI detection. TRIM builds on standard detectors and quantifies feature shifts using prediction entropy and KL divergence.
% Extensive experiments across multiple datasets and attacks show that TRIM achieves 91.97\% and 83.96\% robustness on ProGAN and GenImage datasets, outperforming the state-of-the-art defense by 33.88\% and 28.91\% respectively, while maintaining near original accuracy.
Extensive experiments across multiple datasets and attacks validate the superiority of our TRIM, e.g., outperforming the state-of-the-art defense by 33.88\% (28.91\%) on ProGAN (GenImage), while well maintaining original accuracy.
%As a baseline defense, we first examine adversarial training (AT) and find that it often leads to feature entanglement and eventual performance collapse. Our information-theoretic analysis theoretically and empirically reveals that AT for AIGI detection, fails to preserve feature-label mutual information, thereby rapidly degrades the detector’s discriminative ability. Compared to the feature entanglement caused by AT, we find standard detectors are more susceptible to exhibiting significant feature shifts under adversarial attacks. Leveraging this insight, We propose \textbf{TRIM} (\textbf{T}raining-free \textbf{R}obust Detection via \textbf{I}nformation-theoretic \textbf{M}easures), the first training-free adversarial defense for AIGI detection that quantifies feature shifts via prediction entropy and KL divergence. TRIM is a naturally robust and resilient detector, which can turn standard detectors into adversarial robust ones, without requiring any training. Extensive experiments across multiple datasets and attacks show that TRIM achieves 91.97\% and 83.96\% robustness on ProGAN and GenImage datasets, outperforming the state-of-the-art defense by 37\% and 28.91\% respectively, while maintaining near original accuracy.

%effectively defends against a wide range of attacks while maintaining high accuracy on clean inputs, outperforming state-of-the-art defense methods, including both AT and test-time defenses.
\end{abstract}

% essentially manifests as a degradation of mutual information between feature representations and labels. To further investigate the underlying cause of this failure, we then decouple the optimization objective of AT from an information-theoretic perspective, revealing its limitations in AIGI detection tasks. 

\section{Introduction}
% Recent advances in generative models, such as GANs and Diffusion Models, have enabled AI-generated images (AIGI) to achieve impressive levels of photorealism and customizability, driving their adoption in fields like creative design and virtual reality. However, this progress has also raised serious concerns: synthetic images can be weaponized to spread misinformation, manipulate public opinion, and facilitate copyright infringement or identity forgery. Furthermore, the widespread use of AIGI risks contaminating real-world datasets, undermining the integrity of future model training. These challenges underscore the pressing need for reliable methods to verify the authenticity of visual content.

% In response to the increasing risks posed by AI-generated content, the detection of AI-generated images (AIGI) has gained significant attention. Current methods typically differentiate between real and synthetic images by analyzing artifacts in the spatial or frequency domain, reconstruction errors, and irregular noise patterns. While these approaches have demonstrated high accuracy across various AIGI datasets, previous studies have highlighted their vulnerability to adversarial attacks.Even subtle,imperceptible perturbations can drastically reduce detection accuracy. Recently, novel attack methods targeting both face forgery and AIGI detection have emerged, generating adversarial samples by minimizing statistical or spectrum discrepancies between real and synthetic images, thereby posing new challenges to the adversarial robustness of AIGI detection systems.
AI-generated images (AIGI) achieve great photorealism and customization, sparking urgent demands for authenticity verification due to risks including misinformation propagation, copyright violations, and dataset contamination. Current AIGI detection methods mainly rely on identifying spatial/frequency artifacts\cite{wang2020cnn,liu2020global,zhang2019detecting,frank2020leveraging}, reconstruction errors\cite{zhang2023diffusion,ricker2024aeroblade}, and anomalous noise patterns\cite{tan2023learning,tan2024rethinking,liu2022detecting,zhong2023patchcraft}, etc., achieving high accuracy across synthetic datasets. However, these approaches exhibit critical vulnerabilities to adversarial attacks that evade detection by adding tiny perturbations to synthesized images~\cite{diao2024vulnerabilities,de2024exploring,mavali2024fake,saberi2023robustness,dong2022think,hou2023evading,jia2022exploring,zhou2024stealthdiffusion}.
% Furthermore, emerging adversaries exploit statistical or spectral alignment between real and synthetic distributions, crafting adversarial samples that bypass detection by minimizing domain discrepancies\cite{dong2022think,hou2023evading,jia2022exploring,zhou2024stealthdiffusion}, thereby posing new challenges to the adversarial robustness of AIGI detection systems. 

\begin{figure}[!tbp]
    \centering
    \begin{subfigure}[t]{0.47\textwidth}
        \includegraphics[width=\textwidth]{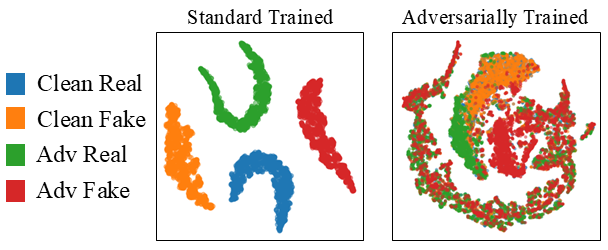}
        \caption{t-SNE visualization of features for standard and adversarially trained detectors, showing feature entanglement in the adversarially trained one.}
        \label{fig:differ_sub1}
    \end{subfigure}
    \hspace{0.8em} 
    \begin{subfigure}[t]{0.48\textwidth}
        \includegraphics[width=\textwidth]{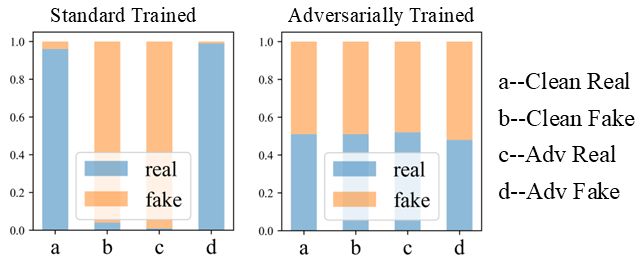}
        \caption{Confidence distribution over real and fake classes (summing to 1), where the adversarially trained detector assigns nearly identical scores to both classes.}
        \label{fig:differ_sub2}
    \end{subfigure}
    % \vspace{-1em}
    \caption{Comparison of standard and adversarially trained detectors.}
    \label{fig:difference}  
    \vspace{-1.5em}
\end{figure}

%While AIGI detectors have been shown to be vulnerable to adversarial attacks, research on effective defense strategies remains limited. \cite{mavali2024fake} (arXiv) improves robustness via unsupervised adversarial fine-tuning on CLIP-based\cite{radford2021learningtransferablevisualmodels} detectors, but is constrained by model architecture and the need for additional adversarial data. To fill this gap, we propose \textbf{TRIM} (\textbf{T}raining-free \textbf{R}obustification via \textbf{I}nformation-theoretic \textbf{M}easures), a universal training-free defense framework based on an information-theoretic perspective. TRIM can turn standard detectors into resilient ones to largely improve adversarial robustness while preserving high accuracy on clean inputs, without the need for any training. To our best knowledge, this is the first training-free defense for AIGI detection.
Nonetheless, defenses have been rarely explored.
Since adversarial training (AT) is among the most effective and widely adopted defenses, we initially apply AT~\cite{madry2019deeplearningmodelsresistant, zhang2019theoreticallyprincipledtradeoffrobustness} to improve the robustness of AIGI detectors. However, it often results in performance collapse, where the loss fails to converge and the detector loses its discriminative capability. Specifically, we observe two symptoms in the latent feature distributions under AT (see \cref{fig:difference}): (1) entangled feature representations between clean and adversarial samples; (2) reduced separability in output confidence scores across classes. From an information-theoretic perspective, these phenomena suggest a decrease in the mutual information (MI) between feature representations and target labels~\cite{tishby2000informationbottleneckmethod}. To gain deeper insight, we reinterpret AT through the lens of MI and decompose its optimization objective into two components: one associated with clean samples and the other with adversarial perturbations. By tracking these two components during training, we observe a trade-off that causes the performance collapse (\cref{fig:AT_MI})---maximizing MI for adversarial samples often comes at the cost of a sharp drop MI for clean samples, which reveals a fundamental limitation of adversarial training in AIGI detection task. 

% Unlike the ambiguous feature distributions on the adversarially trained detector, on the standard trained detector, there is a clear feature shift between misclassified adversarial samples and their corresponding clean samples(see \cref{fig:differ}), while adversarial samples that fail to attack remain within the feature distribution of clean samples. Since clean-adversarial sample pairs are unavailable in practical detection scenarios,we use prediction entropy to reflect the impact of the different feature distribution on the uncertainty of the detector's output, and employ it as a metric to detect misclassified adversarial samples, flipping their labels to obtain the correct output.

\begin{figure}[!tbp]
\centering
\includegraphics[width=\linewidth]{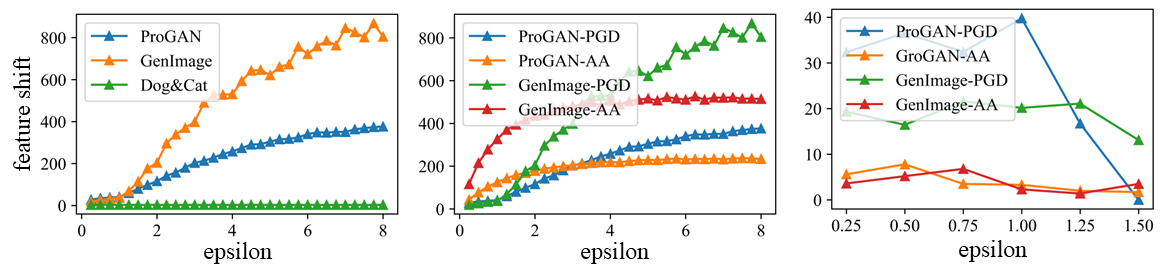}
\vspace{-1em}
\caption{The $l_2$ norm of feature shifts under different adversarial perturbation amplitudes. Left: all PGD adversarial samples. Middle: only successful PGD and AutoAttack samples. Right: only failed adversarial samples. Note that the x-axis in the right plot is set up to $\epsilon \leq 1.5/255$ to ensure there are failed attacks.}
\label{fig:feature_shift}
\vspace{-1.5em}
\end{figure}

Distinct from the feature entanglement induced by AT, standard detectors exhibit a significant deviation in the feature space of misclassified adversarial samples relative to their original clean counterparts, as shown in \cref{fig:feature_shift}. We refer to this phenomenon as \textit{adversarial feature shift}. This observation motivates us to exploit such feature shifts as a signal for identifying adversarial examples. However, since directly measuring these shifts in real-world scenarios is impractical, we establish a theoretical link between adversarial feature shift and information-theoretic measure. By identifying misclassified adversarial samples with informational anomaly metrics, we correct erroneous predictions to enhance robustness. Formally, we propose \textbf{TRIM} (\textbf{T}raining-free \textbf{R}obust Detection via \textbf{I}nformation-theoretic \textbf{M}easures), a training-free defense framework based on an information-theoretic perspective. TRIM is a naturally robust and resilient detector that turns standard pre-trained detectors into robust ones, without requiring any training. To our best knowledge, this is the first training-free defense for AIGI detection. In sum, our contributions are as follows.
% To further counter adaptive attacks that bypass entropy-based detection, we propose an information-theoretic estimation method based on adversarial denoising. By calculating the difference between samples before and after denoising, and using Kullback-Leibler (KL) divergence as an upper bound to quantify the MI between this difference and the detector's output, we can effectively amplify the disparity between misclassified adversarial samples and the other. This significantly enhances detection performance, particularly when prediction entropy fails. Finally, we combine both entropy-based and KL divergence-based detection methods into a two-stage detection system to defend against various adversarial attacks.
\begin{itemize}
\item For the AIGI detection task, we theoretically and empirically uncover the previously unexplored cause of performance collapse in AT through an information-theoretic framework, suggesting potential directions for improvement.

% Through theoretical analysis and empirical validation, we identify and characterize the previously unexplored cause of performance collapse in AT for AIGI detection from an information-theoretic perspective, suggesting potential directions for improvement.
%illustrate the limitations of adversarial training in AIGI detection tasks from the perspective of information theory, suggesting potential directions for improvement.
\item We propose TRIM, the first training-free adversarial defense for AIGI detection, which turns standard detectors into robust ones without requiring additional training.
%universal adversarial defense method for AIGI detection, which can be integrated into detectors as a plug-and-play module without requiring training.
\item Extensive experiments across multiple AIGI datasets, detectors, and adversaries show that TRIM substantially outperforms state-of-the-art defense methods, including both AT and test-time defenses, while preserving near original accuracy.
\end{itemize}
% \begin{figure}[tbp]
%     \centering
%     \includegraphics[width=1\linewidth]{pictures/DIfference.png}
%     \caption{Comparison between normally trained models and adversarially trained models. The first row shows CNNSpot normally trained on ProGAN; the second row shows its adversarially trained version. The first column presents the feature distribution visualized by t-SNE (the high-dimensional inputs to the final classification layer), the second column shows the confidence distribution, and the third column displays the gradient saliency maps obtained through guided backpropagation.}
%     \label{fig:differ}
%     \vspace{-0.8em}
% \end{figure}

%\section{Methodology}

\section{Why Adversarial Training Fails for AI-Generated Image Detection}
\noindent \textbf{Notation.} The mutual information $I(A;B)$ between two random variables $A$ and $B$ quantifies how much knowing one of the variables reduces uncertainty about the other. It is mathematically defined as $I(A;B)=H(A)-H(A|B)=H(B)-H(B|A)$, where $H(A)$ denotes the entropy of $A$, and $H(A|B)$ denotes the conditional entropy of $A$ given $B$.

To investigate the challenges of improving adversarial robustness of AIGI detectorts, we begin by examining adversarially training for AIGI detectors, as AT is widely regarded as the most effective defense against adversarial attacks~\cite{robustbench}. To this end, we conduct extensive evaluation of multiple AT methods on 2 popular AIGI detection datasets: ProGAN\cite{wang2020cnn} and GenImage\cite{zhu2023genimage}. Experimental results in~\cref{tab:main} demonstrate that, regardless of the AT methods employed, models trained on AIGI datasets consistently suffer from the performance collapse. This manifests as non-convergence during training (see Appendix C for details), with both accuracy and robustness stagnating around 50\%, or even below 50\%, matching random guess performance in binary classification. These phenomena are consistently observed across all evaluated detectors~\cite{wang2020cnn,frank2020leveraging,tan2024rethinking,zhong2023patchcraft,ojha2024universalfakeimagedetectors,liu2020global}, breaking the common belief that AT can improve adversarial robustness. Most recently, \cite{diao2024vulnerabilities} also observed performance collapse following AT for AIGI detection. However, to the best of our knowledge, the underlying causes of this failure remain unexplored.
%We evaluate two adversarial training methods, PGD-AT\cite{madry2019deeplearningmodelsresistant} and TRADES\cite{zhang2019theoreticallyprincipledtradeoffrobustness}, on three commonly used AIGI datasets: ProGAN\cite{wang2020cnn}, GenImage\cite{zhu2023genimage}, and Stable Diffusion v1.4(SDv1.4)\cite{rombach2022highresolutionimagesynthesislatent} from GenImage.

% Experiments show that on ProGAN, both PGD-AT and TRADES lead to performance collapse (clean/attack accuracy $\leq$ 50\%, perturbation amplitude adjustment ineffective); on GenImage, PGD-AT collapses while TRADES converges (PGD attack robustness 65\% for CNNSpot); on SDv1.4, PGD-AT achieves 71.14\% robustness for CNNSpot under 8/255 PGD attacks, while TRADES only reaches 53.46\%.

To investigate the underlying causes, we first visualize the latent feature distributions of clean and adversarial samples under standard/adversarially trained detector, and analyze the detector's output confidence across classes. \cref{fig:differ_sub1} shows that, on adversarially trained detector, the latent features of clean and adversarial samples---regardless of being real or fake---to become entangled. Meanwhile, \cref{fig:differ_sub2} shows that the adversarially trained detector assigns nearly identical confidence scores to real and fake classes, revealing poor discriminative ability. In contrast, a standard detector trained only on clean data yields well-separated features and distinct confidence distributions, confirming that the degradation arises from adversarial training. From an information-theoretic perspective, this implies a decrease in the MI between the learned feature representations and the target labels~\cite{tishby2000informationbottleneckmethod}, thereby compromising model’s ability to distinguish fake/real samples. To validate this hypothesis, we quantitatively assess the MI under AT regimes in the next subsection.  

%after adversarial training, the features of clean and adversarial samples, whether real or fake, become heavily entangled, and the confidence across classes are similar in~\cref{fig:differ_sub2}, showing little discrimination. In contrast, a standard detector trained solely on clean data produces well-separated features and distinct class confidences, confirming that the observed degradation stems from adversarial training. 
% These results suggest that adversarial training reduces the detector's ability to effectively represent information, aligning input data with target labels. This issue is central to the Information Bottleneck (IB) principle\cite{tishby2000informationbottleneckmethod}, which views model training as compressing input information into feature representations that maximally preserve label-relevant information. Motivated by this perspective, we further analyze adversarial training through the lens of IB.

%are overlapping and class confidence is indistinguishable, it implies that the information about the label embedded in the features is significantly reduced. In information-theoretic terms, this corresponds to a decrease in MI between the learned feature representations and the target labels.Therefore, we hypothesize that adversarial training suppresses the MI between features and labels, which in turn undermines the model’s ability to distinguish fake from real samples. To validate this hypothesis, we further quantify the MI under adversarial training regimes.
\begin{figure}[!tbp]
  \centering
  \includegraphics[width=1\linewidth]{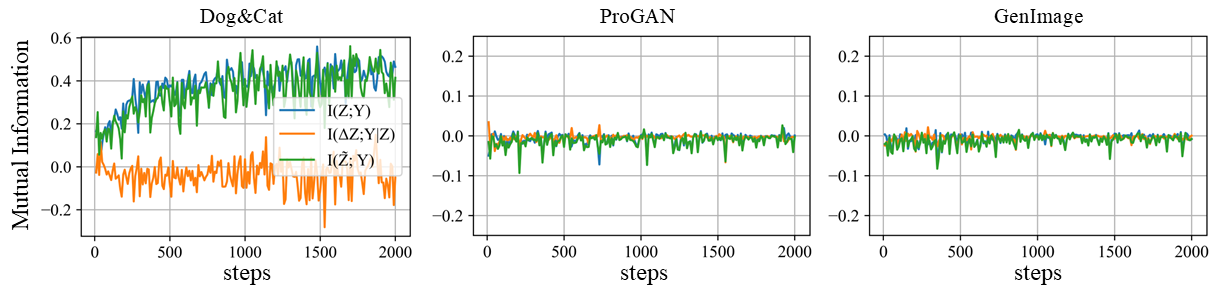} 
  \vspace{-1em}
  \caption{Mutual information $I(Z; Y)$, $I(\Delta Z; Y|Z)$ and $I(\tilde Z; Y)$ as a function of training steps.}
  \label{fig:AT_MI} 
\vspace{-1.5em}
\end{figure}

\noindent\textbf{Rethinking Adversarial Training via Mutual Information.}
From an information-theoretic perspective, the optimization objective of standard training can be formulated via MI:
\begin{equation}
    \underset{\theta}{\max}\, I(Z; Y), 
    \label{eq:standard_MI}
\end{equation}
where $Z\in R^{d}$ is the compressed feature of the input image $X$, and $Y$ is its corresponding label. $I(\cdot;\cdot)$ denotes the MI between two variables. $I(Z; Y)$ quantifies how much information $Z$ retains about $Y$, is decomposed into $H(Y)-H(Y|Z)$, $H(Y)$ describes the distribution of labels in the dataset, while $H(Y|Z)$ describes the uncertainty of predicting $Y$ based on $Z$. $H(Y|Z)$ is usually estimated through cross entropy\cite{achille2018emergence}\cite{achille2018information}\cite{amjad2019learning} and is therefore optimized with cross-entropy loss. In AT, let $\tilde{Z}$ denote the feature representation of adversarial inputs, the optimization objective of AT can be reinterpreted via MI:
% The MI $I(A;B)$ between two random variables $A$ and $B$ quantifies how much knowing one of the variables reduces uncertainty about the other, it is mathematically defined as
% \begin{equation}
%    I(A;B)=H(A)-H(A|B)=H(B)-H(B|A),
%     \label{eq:MI}
% \end{equation}
% where $H(A)$ is the entropy of $A$, and $H(A|B)$ is the conditional entropy of $A$ given $B$.During training, $I(Z;Y)$, which quantifies how much information $Z$ retains about $Y$, is decomposed into $H(Y)-H(Y|Z)$, $H(Y)$ describes the distribution of labels in the dataset,while $H(Y|Z)$ describes the uncertainty of predicting $Y$ based on $Z$. $H(Y|Z)$ is usually estimated through cross entropy\cite{achille2018emergence}\cite{achille2018information}\cite{amjad2019learning} and is therefore optimized with cross-entropy loss. 
\begin{equation}
    \underset{\theta}{\max}\, I(\tilde{Z};Y)
    \label{eq:AT_MI}.
\end{equation}
Next, we provide Proposition 1 to disentangle $I(\tilde{Z};Y)$ and illustrate the transformation relationship of MI among the four variables.

\textbf{Proposition 1.} Let $ Z, \tilde{Z}, \Delta Z, Y $ denote four random variables, where $\tilde{Z}=Z+\Delta Z$. Then the following approximate relationship holds:
\begin{equation}
    I(\tilde{Z};Y)\approx I(Z;Y)+I(\Delta Z;Y|Z).
\end{equation}
A detailed proof is provided in Appendix B. Consequently, $I(\tilde{Z}; Y)$ can be naturally decomposed into two competing objectives: (1) $I(Z; Y)$, which measures the model’s predictive performance on clean inputs, and (2) $I(\Delta Z; Y | Z)$, which captures the conditional dependency between adversarial perturbation information and labels given its corresponding  original features.

To investigate the dynamic behavior of the two optimization objectives during AT, we train ResNet-50 on two binary classification tasks: AIGI detection and Dog-vs-Cat classification\cite{dogs_vs_cats}. We record the values of $I(Z; Y)$, $I(\Delta Z; Y |Z)$ and $I(\tilde Z; Y)$ at each training steps (see Appendix C for detailed computation of MI). As shown in \cref{fig:AT_MI}, On both the Dog-vs-Cat and AIGI datasets, $I(\Delta Z; Y| Z)$ remains low, which suggests that AT forces the adversarial perturbation to behave more like random noise that is independent of $Y$. Furthermore, on AIGI datasets, $I(Z; Y)$ does not show any upward trend and instead fluctuates around zero. Consequently, as the sum of the two MI, $I(\tilde{Z}; Y)$ also stays minimal. By analyzing the results, we have two findings: (1) the performance collapse after AT is unique to AIGI detetcion, not on general binary object recognition; (2) For AIGI detection tasks, AT often leads to a reduction of MI for clean samples, $I(Z; Y)$. This weakens the model's discriminative ability, resulting in feature entanglement and eventual performance collapse. Therefore, it is essential to develop a new adversarial defense strategy for AIGI detection to replace AT.

\section{Training-free Robust Detection via Information-theoretic Measures}
\subsection{Training-free Robust Detection Based on Predictive Entropy }
\label{section_entropy}

% The above analysis indicates that the significant informational conflict between feature shift ($\Delta Z$) and clean features ($Z$) undermines the effectiveness of adversarial training on the AIGI detector. Meanwhile, recent advances have introduced training-free detection methods that distinguish real and fake images via score functions, however, these approaches are incompatible with adversarial training. In light of this, we shift our focus in this section to examining the feature behavior of original models considering only clean samples, aiming to identify adversarial samples that mislead the detector and subsequently performing label correction.Under the framework of MI theory, we analyze the informational dependency between internal features and model outputs to investigate the detectability and correctability of adversarial samples.

Although the feature distributions of adversarially trained models are highly entangled, we observe that, under standard training, clean samples and their adversarial counterparts exhibit significant separations in feature space(see~\cref{fig:differ_sub1}) due to the notable shifts in feature representations caused by adversarial perturbation. We refer to this phenomenon as \textit{adversarial feature shift}. To further validate this observation, we measure the adversarial feature shift on AIGI detection dataset and Dog-vs-Cat dataset respectively. Specifically, we train ResNet-50 on ProGAN, GenImage and Dog-vs-Cat datasets as the victim models, apply PGD attacks with varying perturbation budgets (ranging from $1/255$ to $8/255$) , and plot the relationship between the $l_2$ norm of $\Delta Z$ and the perturbation magnitude in \cref{fig:feature_shift}. \cref{fig:feature_shift} (left) shows that the AIGI dataset consistently exhibits larger feature shifts than the Dog-vs-Cat dataset, even under small perturbations. Notably, these shifts are primarily driven by successful adversarial samples, which induce substantial changes in feature space (\cref{fig:feature_shift} (middle)), whereas failed adversarial samples (i.e., still classified correctly) result in much smaller shifts (\cref{fig:feature_shift} (right)). Therefore, we argue that adversarial feature shift is a key insight for identifying misclassified adversarial samples, which naturally raises the question: \textit{For AIGI detection, how can one make use of the property of adversarial feature shifts to develop a new defense mechanism?}

Unfortunately, in real-world scenarios, it is impractical to explicitly measure the adversarial feature shift because the original (pre-attack) images of the adversarial inputs are unknown. To tackle this challenge, we first build the connection between adversarial feature shift and MI via Proposition 2.

\textbf{Proposition 2.}
Let $\tilde{y}$ denote the softmax output of an AIGI detector and $\tilde{Z}$ the adversarial feature. 
\begin{itemize}
\item For successful adversarial samples(misclassified), the model's prediction is mainly influenced by the feature shift $\Delta Z$  caused by adversarial perturbation, i.e.,
 $I(\tilde{Z}; \tilde{y}) \approx I(\Delta Z; \tilde{y})$.
 
Using $I(A; B) = H(B) - H(B|A)$, we have:
\begin{equation}
H(\tilde{y} | \tilde{Z}) \approx H(\tilde{y} | \Delta Z) .
\label{eq:entropy_success}
\end{equation}
\item For unsuccessful adversarial samples(correctly classified), the prediction remains largely governed by the original clean feature $Z$ , i.e., $I(\tilde{Z}; \tilde{y}) \approx I(Z; \tilde{y})$, which implies:
\begin{equation}
H(\tilde{y} | \tilde{Z}) \approx H(\tilde{y} | Z ).
\label{eq:entropy_fail}
\end{equation}
\end{itemize}
A detailed proof is provided in Appendix B. The prediction entropy $H(\tilde{y} | \tilde{Z})$ for a single sample is computed as:
\begin{equation}
H(\tilde{y} | \tilde{Z}) = -\sum_{y_i \in \mathcal{Y}} p(y_i | \tilde{Z}) \log p(y_i  | \tilde{Z}),
\end{equation}
where $\mathcal{Y}$ denotes the set of all possible labels in the dataset and the probabilities are obtained from the model’s softmax output.

\begin{figure}[!tbp]
    \centering
    \begin{subfigure}[t]{0.48\textwidth}
        \includegraphics[width=\textwidth]{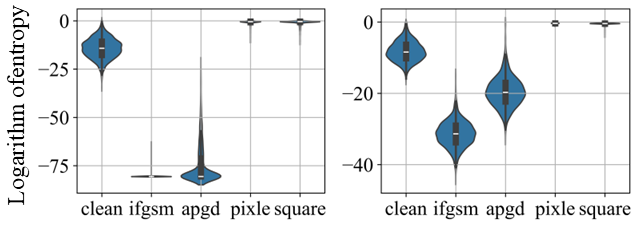}
        \caption{Logarithm of predictive entropy of successful adversarial samples relative to the predicted entropy of clean samples. The left column is CNNSpot, and the right column is UnivFD. }
        \label{fig:entropy_sub1}
    \end{subfigure}
    \hspace{0.8em} 
    \begin{subfigure}[t]{0.48\textwidth}
        \includegraphics[width=\textwidth]{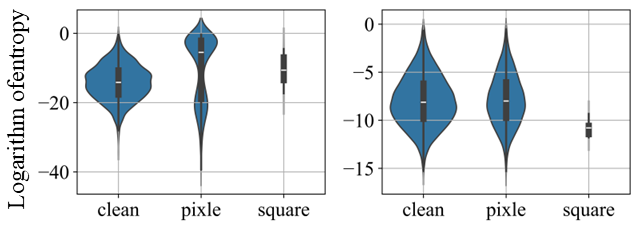}
        \caption{Logarithm of predictive entropy of unsuccessful adversarial samples. Since all adversarial samples generated by APGD and IFGSM are misclassified, they are not included in the subgraph.}
        \label{fig:entropy_sub2}
    \end{subfigure}
    % \vspace{-1em}
    \caption{Distribution of predicted entropy of adversarial samples relative to clean samples.}
    \label{fig:entropy}
    \vspace{-1em}
\end{figure}

\begin{figure}[!tbp]
  \centering
\includegraphics[width=\linewidth]{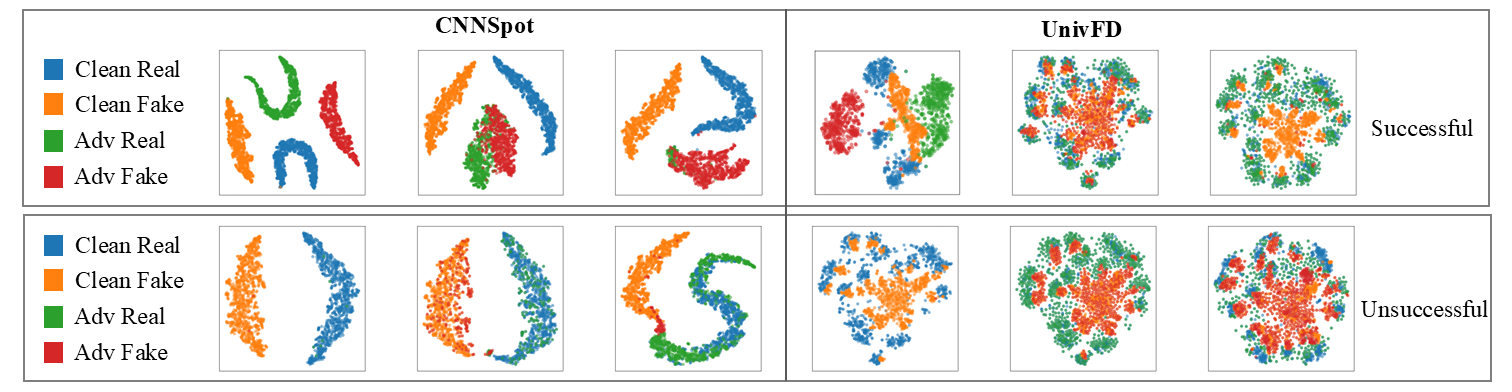} 
\vspace{-1em}
  \caption{Feature distributions of adversarial and clean samples under various attacks on CNNSpot and UnivFD.}
  \label{fig:feature_tsne}
  \vspace{-1.5em}
\end{figure}

According to \cref{eq:entropy_success} and \cref{eq:entropy_fail}, we model the relationship between adversarial feature shifts and the predictive entropy of adversarial samples. Instead of directly measuring feature shifts, we compute the prediction entropy and visualize the results in \cref{fig:entropy}. Interestingly, successfully misclassified adversarial samples exhibit opposite entropy behaviors depending on the attack type: white-box attacks (e.g., I-FGSM~\cite{madry2019deeplearningmodelsresistant}, APGD~\cite{croce2020reliableevaluationadversarialrobustness}) tend to yield low entropy, while black-box attacks (e.g., Pixle~\cite{pixle}, Square~\cite{andriushchenko2020squareattackqueryefficientblackbox}) often produce high-entropy predictions. To further investigate this contrast, we visualize the adversarial feature distributions on both CNN and Clip based detectors in \cref{fig:feature_tsne}. Under white-box attacks, adversarial real and fake samples form well-separated clusters. In contrast, black-box adversaries produce overlapping distributions. This divergence likely stems from the attack mechanisms: white-box attacks exploit model gradients to confidently suppress true-class probability, whereas black-box methods, lacking gradient access, push inputs toward decision boundaries, resulting in higher uncertainty.

Building on this property and our empirical evidence, we propose a simple yet highly effective, training‑free method for robust AIGI detection. Specifically, we introduce lower and upper bounds on predictive entropy to distinguish adversarial samples: any input with entropy falling outside this range is flagged as a successfully attacked sample. Because real‑vs‑fake detection is a binary classification task, We then invert the detector’s predicted label for each identified adversarial sample, thereby restoring its correct prediction.

%Based on this, we introduce two entropy thresholds (a lower bound and an upper bound) to identify such extreme cases. Samples falling outside this range are considered adversarial samples that have successfully mislead the detector, and they will be re-labeled with the opposite class prediction to counteract their misleading effect. \YF{Move this part to related work:It is worth noting that previous studies\YF{need to add ref} have also explored using uncertainty as a metric for detecting adversarial samples, typically within a Bayesian modeling framework. However, such methods have shown limited effectiveness in AIGI detection tasks. In contrast, TRIM achieves excellent performance in AIGI detection without the need for Bayesian training or fine-tuning, demonstrating its efficiency and effectiveness.}

\subsection{Mitigating Adaptive Attacks via MI with Adversarial Denoising}

Evaluating with adaptive attacks~\cite{tramer2020adaptiveattacksadversarialexample,carlini2017adversarialexampleseasilydetected} is considered essential for a credible defense, as these adversarial samples are specifically customized with full knowledge of the defense mechanism to effectively bypass it. Because our defense mechanism relies on entropy bounds, we design an adaptive attack that forces adversarial samples to exhibit  entropy values similar to benign inputs, which is achieved by constraining the perturbation magnitude or by directly optimizing the logit space(see \cref{sec:adaptive} for detail). In this section, we introduce a mutual information estimation method based on random preprocessing to address this problem.

Based on the observation that adaptive attacks targeting entropy-based detection tend to introduce relatively subtle perturbations, we apply random preprocessing as an effective denoising mechanism to weaken or disrupt these adversarial signals. Specifically, we employ three complementary operations: \textit{GaussianBlur}, \textit{RandomResizedCrop}, and \textit{RandomHorizontalFlip}. GaussianBlur suppresses high-frequency components of the perturbation, while RandomResizedCrop and RandomHorizontalFlip introduce random spatial transformations that break the structured patterns of adversarial noise.

% We first attempted the method proposed in \cite{Xu_2018}, which calculates the L1 norm difference between the detector's output logits before and after filtering to distinguish adversarial samples from clean samples. However, we found that this approach could not effectively separate the two. We also tried directly using the outputs after random preprocessing as the final predictions, but a portion of adversarial samples were still misclassified. Moreover, it is important to note that our goal is only to detect adversarial samples that are misclassified, rather than identifying all adversarial samples. 

We first attempt to use the outputs after denoising as the final predictions, but a portion of adversarial samples are still misclassified. Therefore, instead of relying directly on the denoised results,we leverage adversarial denoising to amplify the differences between clean and adversarial samples, with the goal of detecting adversarial samples based on these differences. 

\begin{wrapfigure}{r}{0.5\textwidth}  
\vspace{-1.5em}
  \centering
  \includegraphics[width=0.49\textwidth]{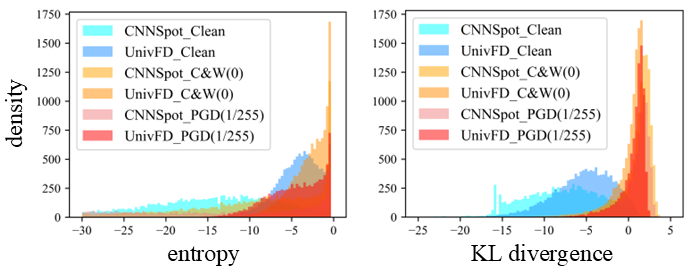} 
  \vspace{-0.5em}
  \caption{Histogram of entropy and KL divergence for clean samples and adversarial examples generated by PGD ($\ell_\infty$, $1/255$) and C\&W ($\ell_2$, $\kappa=0$) attacks on CNNSpot and UnivFD.}
  \label{fig:kl}
  \vspace{-1em}
\end{wrapfigure}

% Based on the above insight, we propose a new method: 
To better capture such differences, we compute the change in feature representations before and after denoising in the feature space, denoted as $Z^b-Z^a$, where $Z^b$ represents the feature before denoising and $Z^a$ the one after denoising. We then estimate $I(Z^b - Z^a; y^b | Z^a)$, the MI between $Z^b-Z^a$ and the detector’s output prior to denoising $y^b$, conditioned on the post-denoising feature $Z^a$. Specifically, we hypothesize that (1) for misclassified adversarial samples, the adversarial information contained in $Z^b-Z^a$ exhibits a strong dependency on the incorrect output $y^b$; (2) In contrast, for correctly classified adversarial samples, the dependency between the adversarial information and the correct output is weaker. (3) Furthermore, for clean samples, $Z^b-Z^a$ carries limited information and shows low correlation with the output $y^b$, resulting in a lower MI value. However, since directly measuring $I(Z^b - Z^a; y^b | Z^a)$ is intractable, we propose Proposition 3 to approximate it by using the Kullback-Leibler (KL) divergence as an upper bound:

\textbf{Proposition 3.} Let $Z^b$ and $Z^a$ denote the feature before and after denoising, respectively, and let $y^b$ and $y^a$ represent the corresponding softmax outputs. Then, the following inequality holds:
\begin{equation} 
I(Z^b - Z^a; y^b | Z^a)  \approx H(y^b | Z^a) - H(y^b | Z^b) \  \leq D_{\text{KL}}(y^b | Z^b, y^a | Z^a),  
\end{equation}
where $D_{\text{KL}}(\cdot, \cdot)$ denotes the KL divergence between the detector outputs before and after adversarial denoising. The inequality becomes an equality if and only if $Z^a = Z^b$ (see Appendix B for detailed proof.) Therefore, we use KL divergence as an indicator to further amplify $I(Z^b - Z^a;y^b|Z^a)$ between clean and adversarial samples. As shown in \cref{fig:kl}, for CNN-based and CLIP-based detectors, adversarial samples that successfully mislead the detectors exhibit higher KL divergence.

Finally, we integrate entropy-based defense with KL divergence-based defense in a two-stage detection framework. The complete procedure is summarized in \cref{alg:trim}. We refer to our method as Training-free Robust Detection via Information-theoretic Measures, or TRIM.

% \subsection{TRIM: Combining Entropy KL divergence Bounds }

% Our proposed TRIM framework combines entropy-based detection and KL divergence through adversarial denoising. The complete procedure is summarized in \cref{alg:trim}:
\begin{algorithm}[htbp]
\small
\caption{TRIM: Training-free Robustification via Information-theoretic Measures}
\label{alg:trim}
\setlength{\baselineskip}{0.95\baselineskip}
\begin{algorithmic}[1]
\REQUIRE Input $X^b$, entropy bounds $[H_{\min}, H_{\max}]$, detector $f_\theta$, KL threshold $\tau$
\ENSURE Corrected prediction $\hat{y}$

\STATE \textbf{Step 1: Entropy-Based Detection and Correction}
\STATE Obtain the softmax output $y^b \leftarrow f_\theta(X^b)$
\STATE Compute entropy $H(y^b|X^b) \leftarrow -\sum_{y_i \in \mathcal{Y}} p(y_i | X^b) \log p(y_i  | X^b)$
\IF{$H(y^b|X^b) \notin [H_{\min}, H_{\max}]$}
    \STATE $\hat{y} \leftarrow 1 - \arg\max p(y^b)$ 
    \RETURN $\hat{y}$
\ENDIF

\STATE \textbf{Step 2: KL-Based Defense via Denoising}
\STATE Generate denoised sample $X^a \leftarrow \text{RandomizedDenoiser}(X^b)$
\STATE Obtain the softmax output $y^a \leftarrow f_\theta(X^a)$
\STATE Compute KL divergence: $D_{KL}(p(y^b | X^b) \parallel p(y^a | X^a))$

\IF{$D_{KL} > \tau$}
    \STATE $\hat{y} \leftarrow 1 - \arg\max p(y^b)$
\ELSE
    \STATE $\hat{y} \leftarrow \arg\max p(y^b)$
\ENDIF

\RETURN $\hat{y}$
\end{algorithmic}
\end{algorithm}
\vspace{-1.5em}

\section{Experiments}
\subsection{Experimental Settings}
\textbf{Datasets.} We evaluate TRIM on three AIGI datasets spanning diverse generative models, including ProGAN~\cite{wang2020cnn}, GenImage~\cite{zhu2023genimage} and SDv1.4~\cite{rombach2022highresolutionimagesynthesislatent}. ProGAN contains 360K real images from LSUN \cite{yu2015lsun} and 360K fake images generated by ProGAN \cite{karras2018progressive}. GenImage is a large-scale benchmark of over 2.6M images, including real images comprising ImageNet real images and fakes from eight advanced GAN-based and Diffusion-based generators. We sample a balanced subset of 360K real and 360K synthetic images for evaluation. SDv1.4 \cite{rombach2022highresolutionimagesynthesislatent} only contains synthetic images generated by Stable Diffusion V1.4~\cite{rombach2022highresolutionimagesynthesislatent} and is used to assess robustness in cross-generator classification. 
\vspace{-0.5em}

\textbf{Evaluated Detectors.} We test TRIM across four representative AIGI detectors, each leveraging different detection cues: CNNSpot\cite{wang2020cnn} learns global artifacts; UnivFD\cite{ojha2024universalfakeimagedetectors} employ a pretrained CLIP to capture semantic differences; FreqNet\cite{tan2024frequencyawaredeepfakedetectionimproving} focuses on high-frequency forgery patterns; and NPR\cite{tan2024rethinking} models local pixel correlations. All are binary classifiers trained with cross-entropy loss.
\vspace{-0.5em}

\textbf{Evaluated Attacks.} We evaluate robustness against 6 adversaries. For white-box attack, we employ PGD~\cite{madry2019deeplearningmodelsresistant}, C\&W~\cite{carlini2017evaluatingrobustnessneuralnetworks} (\(l_2\) norm with \(\kappa = 10\)) and FAB~\cite{croce2020minimallydistortedadversarialexamples}. Stealthdiffusion~\cite{zhou2024stealthdiffusion} is the state-of-the-art attack specifically designed for AIGI detection, so we also include it. For black-box attack, we adopt Square~\cite{andriushchenko2020squareattackqueryefficientblackbox}. AutoAttack~\cite{croce2020reliableevaluationadversarialrobustness} is a strong robust evaluation protocol that ensembles a series white-box and black-box attacks. All non-C\&W attacks use an \(l_\infty\) bound of  \(8/255\).
%During the training protocol of CNNSpot and UnivFD, data augmentation with Blur + JPEG(0.1) is applied. 
%For C\&W attack, we use the \(l_2\) norm with \(\kappa = 10\), while for all other attacks, we set \(8/255\) perturbation bound under the \(l_\infty\) setting.
%under both white-box and black-box settings, covering various attack strategies including PGD~\cite{madry2019deeplearningmodelsresistant}, C\&W~\cite{carlini2017evaluatingrobustnessneuralnetworks}, and AutoAttack~\cite{croce2020reliableevaluationadversarialrobustness}. During the evaluation, we observe that due to the inherent vulnerability of the AIGI detector, AutoAttack achieves a high success rate in its first round using APGD\cite{croce2020reliableevaluationadversarialrobustness}. Therefore, to more objectively reflect the defense performance, we additionally report results for FAB\cite{croce2020minimallydistortedadversarialexamples} and Square Attack\cite{andriushchenko2020squareattackqueryefficientblackbox} with APGD excluded. In addition, We also include the state-of-the-art attack for AIGI detection proposed in \cite{zhou2024stealthdiffusion}, which called Stealthdiffusion. For the C\&W attack, we use the \(l_2\) norm with \(\kappa = 10\), while for all other attacks, the perturbation is constrained under the \(l_\infty\) norm with a maximum magnitude of \(8/255\).
\vspace{-0.5em}

\textbf{Compared Methods.} To the best of our knowledge, TRIM is the first training-free defense for AIGI detection, and thus there is no method for direct comparison. A recent technical report~\cite{mavali2024fake} directly adopt adversarial fine-tuned CLIP encoder~\cite{radford2021learningtransferablevisualmodels} for AIGI detection. We hence use it as a baseline. In addition, we also compare with diverse defense strategies, including AT, adversarial detection and adversarial purification. For AT, we adopt the commonly used PGD-AT and TRADES, as well as the state-of-the-art randomized adversarial training approach (RAT)\cite{jin2023randomizedadversarialtrainingtaylor}. For test-time defense, we choose adversarial detection Feature squeeze~\cite{Xu_2018} and two state-of-the-art adversarial purification method DiffPure~\cite{nie2022diffusionmodelsadversarialpurification} and TPAP~\cite{tang2024robust}.
\vspace{-0.5em}

\textbf{Implementation Details.} TRIM involves two sets of hyperparameters. The first set consists of the upper and lower thresholds required for the entropy-based defense. The second set includes the threshold used for the KL-divergence-based denoising defense(see Appendix C for detailed settings). In addition, the specific configurations for the random preprocessing used during the denoising process are also detailed in Appendix C.

\vspace{-0.5em}
\subsection{Robustness Evaluation against Various Attacks}
As shown in \cref{tab:main}, TRIM outperforms all adversarial defense methods in robustness by a big margin, with only a minor sacrifice in clean accuracy. TRIM achieves an average robustness of \textbf{91.97\%} on the ProGAN dataset and \textbf{83.96\%} on the GenImage dataset, surpassing the 2nd-best defense (TPAP on ProGAN and DiffPure on GenImage) by \textbf{33.88\%} and \textbf{28.91\%}, respectively. This demonstrates that almost all defense methods, either AT or test-time defense, are essentially making random guesses against adversarial examples, failing to provide robustness. The only except is TPAP and Feature Squeeze on special cases. Under the UnivFD + ProGAN setting, TPAP achieves approximately 96\% robustness against Square and StealthDiffusion. However, this robustness is inconsistent, as the method fails to defend against other 46 attack scenarios (attack vs. datasets vs. detectors). Feature Squeeze similarly exhibits inconsistency issues, demonstrating robustness only in limited scenarios. More critically, research has shown that Feature Squeeze cannot defend against adaptive attacks\cite{Xu_2018}, and thus cannot be considered a reliable defense method. In contrast, TRIM demonstrates consistent superiority in 46 out of all 48 scenarios.

\begin{table}[tbp]
  \centering
  \caption{Evaluation of accuracy(\%) of clean samples and robustness(\%) under different attacks. In the attacks, ``AA'' denotes AutoAttack and ``Sdiff'' denotes StealthDiffusion. In the baselines, ``Original'' refers to the standard trained model without any defense mechanism and ``Fsqueeze'' denotes Feature Squeeze.``Ave''denotes the average robustness against all attacks.}
  \resizebox{1.0\linewidth}{!}{
    \begin{tabular}{cc|c|cccccc|c|c|cccccc|c}
    \hline
    \multirow{3}{*}{Detctors} & \multirow{3}{*}{Baselines} & \multicolumn{8}{c|}{Dataset: ProGAN}                          & \multicolumn{8}{c}{Dataset: GenImage} \\
\cline{3-18}          &       & \multirow{2}{*}{Clean} & \multicolumn{6}{c|}{Atack}                    & \multirow{2}{*}{Ave} & \multirow{2}{*}{Clean} & \multicolumn{6}{c|}{Attack}                   & \multirow{2}{*}{Ave} \\
\cline{4-9}\cline{12-17}          &       &       & PGD   & C\&W  & AA    & FAB   & Square & Sdiff &       &       & PGD   & C\&W  & AA    & FAB   & Square & Sdiff &  \\
    \hline
    \multirow{8}[4]{*}{CNNSpot} & Original & 99.89 & 0     & 0     & 0     & 16.85 & 0.2   & 1.75  & 3.13  & 95.49 & 0     & 0     & 0     & 18.88 & 0.94  & 0.87  & 3.45  \\
\cline{2-18}          & PGD-AT & 50.63 & 49.9  & 50.58 & 44.18 & 50.6  & 45.39 & 49.86 & 48.42  & 50.67 & 49.72 & 50.57 & 49.72 & 49.87 & 50.99 & 50.14 & 50.17  \\
          & TRADES & 50.12 & 6.54  & 50.1  & 33.03 & 50.12 & 50.09 & 12.32 & 33.70  & 55.22 & 9.41  & 44.62 & 9.41  & 25.88 & 40.59 & 9.46  & 23.23  \\
          & RAT   & 50.17 & 0.9   & 49.97 & 3.27  & 29.84 & 4.93  & 8.57  & 16.25  & 50.02 & 0.74  & 47.77 & 9.85  & 49.37 & 19.86 & 1.13  & 21.45  \\
          & Fsqueeze & 98.35 & 97.95 & 4.18  & 0.44  & 18.78 & 0.44  & 96.24 & 36.34  & 83.1  & 100   & 18.78 & 2.06  & 19.68 & 2.08  & 98.77 & 40.23  \\
          & DiffPure & 53.37 & 53.96 & 53.96 & 53.7  & 53.96 & 53.95 & 53.89 & 53.90  & 66.86 & 59.52 & 61.9  & 60.84 & 61.89 & 61.9  & 60.05 & 61.02  \\
          & TPAP  & 52.78 & 51.42 & 50.93 & 48.71 & 52.68 & 49.35 & 51.64 & 50.79  & 52.27 & 50.6  & 52.49 & 49.83 & 52.27 & 51.14 & 50.8  & 51.19  \\
          & TRIM  & \textbf{98.9} & \textbf{100} & \textbf{82.56} & \textbf{90.74} & \textbf{99.83} & \textbf{91.46} & \textbf{99.8} & \textbf{94.57 } & \textbf{89.56} & \textbf{100} & \textbf{73.21} & \textbf{74.01} & \textbf{96.37} & \textbf{77.03} & \textbf{99.83} & \textbf{86.74 } \\
    \hline
    \multirow{8}[4]{*}{UnivFD} & Original & 98.76 & 0     & 0.05  & 0     & 58.61 & 0.04  & 1.58  & 10.05  & 91.79 & 0     & 0.34  & 0     & 46.06 & 0     & 1.79  & 8.03  \\
\cline{2-18}          & PGD-AT & 69.61 & 0     & 41.58 & 0     & 36.15 & 0.48  & 0     & 13.04  & 49.73 & 0.68  & 49.43 & 23.05 & 49.63 & 46.25 & 2.63  & 28.61  \\
          & TRADES & 95.49 & 0     & 72.27 & 0.06  & 51.32 & 0.39  & 0.07  & 20.69  & 56.81 & 0.04  & 52.8  & 6.83  & 50.31 & 40.04 & 3.4   & 25.57  \\
          & RAT   & 98.32 & 0.2   & 0.85  & 0.13  & 48.65 & 0.76  & 0.22  & 8.47  & 51.78 & 15.94 & 51.78 & 19.85 & 50.62 & 50.74 & 17.26 & 34.37  \\
          & Fsqueeze & \textbf{98.58} & 32.68 & 31.16 & 0.04  & 58.61 & 0.04  & 29.84 & 25.40  & \textbf{90.79} & 0.85  & 1.87  & 0     & 46.01 & 0     & 0.91  & 8.27  \\
          & DiffPure & 62.5  & 60.71 & 62.49 & 60.3  & 62.48 & 61.5  & 60.97 & 61.41  & 60.71 & 59.63 & 56.72 & 56.2  & 57.31 & 55.76 & 58.86 & 57.41  \\
          & TPAP  & 71.3  & 98.74 & 58.17 & 56.32 & 60.96 & \textbf{96.45} & \textbf{96.3} & 77.82  & 51.94 & 47.73 & 50.6  & 47.7  & 50.16 & 52.75 & 47.23 & 49.36  \\
          & TRIM  & 98.18 & \textbf{100} & \textbf{93.88} & \textbf{87.62} & \textbf{98.77} & 91.82 & 96.13 & \textbf{94.70 } & 85.19 & \textbf{100} & \textbf{60.3} & \textbf{71.97} & \textbf{92.42} & \textbf{82.81} & \textbf{95.75} & \textbf{83.88 } \\
    \hline
    \multirow{8}[4]{*}{FreqNet} & Original & 95.78 & 0.09  & 0.14  & 0     & 95.78 & 1.13  & 1.49  & 16.44  & 95.83 & 0.09  & 0.04  & 0.09  & 94.42 & 0.78  & 0.56  & 16.00  \\
\cline{2-18}          & PGD-AT & 57.82 & 45.27 & 52.23 & 28.98 & 57.82 & 29.15 & 46.98 & 43.41  & 50    & 42.67 & 48.27 & 25.32 & 42.32 & 29.9  & 42.64 & 38.52  \\
          & TRADES & 50.63 & 8.36  & 47.79 & 9.1   & 50.63 & 11.51 & 8.92  & 22.72  & 68.5  & 17.17 & 47.04 & 20.81 & 68.5  & 21.64 & 21.78 & 32.82  \\
          & RAT   & 50.69 & 0.49  & 0.59  & 2.1   & 50.69 & 6.58  & 1.49  & 10.32  & 66.17 & 24.82 & 35.21 & 24.36 & 64.58 & 41.17 & 24.51 & 35.78  \\
          & Fsqueeze & 90.25 & 83.85 & 18.78 & 7.33  & 90.25 & 9.84  & 86.37 & 49.40  & \textbf{95.17} & 60.08 & 0.34  & 22.24 & 95.47 & 15.05 & 58.39 & 41.93  \\
          & DiffPure & 49.8  & 53.57 & 50.24 & 50.38 & 50.12 & 49.4  & 52.65 & 51.06  & 50.59 & 51.19 & 51.16 & \textbf{51.19} & 50.62 & 51.28 & 51.25 & 51.12  \\
          & TPAP  & 52.77 & 52.1  & 52.53 & 51.62 & 52.7  & 52.48 & 52.25 & 52.28  & 51.83 & 50.57 & 51.09 & 49.86 & 51.8  & 50.14 & 50.2  & 50.61  \\
          & TRIM  & \textbf{93.47} & \textbf{95.05} & \textbf{73.54} & \textbf{60.95} & \textbf{93.47} & \textbf{69.84} & \textbf{93.51} & \textbf{81.06 } & 85.72 & \textbf{83.02} & \textbf{73.08} & 48.03 & \textbf{85.72} & \textbf{79.03} & \textbf{61.58} & \textbf{71.74 } \\
    \hline
    \multirow{8}[4]{*}{NPR} & Original & 95.68 & 0.15  & 0.3   & 0     & 50.74 & 0.05  & 0.22  & 8.58  & 95.44 & 0.29  & 0     & 0.38  & 39.09 & 0.83  & 0.74  & 6.89  \\
\cline{2-18}          & PGD-AT & 51.5  & 51.5  & 51.5  & 48.48 & 51.5  & 49.43 & 50.86 & 50.55  & 61.73 & 52.48 & 60.27 & 52.48 & 58.03 & 55.17 & 53.16 & 55.27  \\
          & TRADES & 51.5  & 16.16 & 1.89  & 23.22 & 42.6  & 49.87 & 21.52 & 25.88  & 51.53 & 50.58 & 51.31 & 47.26 & 49.61 & 53.32 & 51.75 & 50.64  \\
          & RAT   & 50.84 & 0.41  & 50.58 & 1.32  & 11.71 & 11.48 & 0.94  & 12.74  & 60.07 & 19.47 & 59.88 & 29.64 & 47.46 & 47.78 & 26.77 & 38.50  \\
          & Fsqueeze & 81.27 & 63.22 & 4.5   & 0.4   & 41.99 & 0.4   & 60.39 & 28.48  & 72.78 & 52.14 & 20.3  & 25.8  & 27.55 & 73.5  & 52.58 & 41.98  \\
          & DiffPure & 53.57 & 54.16 & 53.58 & 52.62 & 53.51 & 53.17 & 53.92 & 53.49  & 50.79 & 51.36 & 50.25 & 50.6  & 50.14 & 50.59 & \multicolumn{1}{r|}{50.9} & 50.64  \\
          & TPAP  & 51.72 & 51.72 & 51.77 & 50.72 & 51.3  & 51.69 & 51.7  & 51.48  & 55.92 & 54.17 & 56.43 & 54.08 & 55.27 & 54.26 & 54.31 & 54.75  \\
          & TRIM  & \textbf{92.22} & \textbf{99.98} & \textbf{94.78} & \textbf{95.65} & \textbf{96.31} & \textbf{99.38} & \textbf{99.25} & \textbf{97.56 } & \textbf{92.01} & \textbf{99.65} & \textbf{75.94} & \textbf{90.11} & \textbf{97.07} & \textbf{98.34} & \textbf{99.73} & \textbf{93.47 } \\
    \hline

    \end{tabular}
    }
  \label{tab:main}
  \vspace{-0.8em}
\end{table}

\noindent \textbf{Comparison with UnivFD-AF.}
UnivFD-AF~\cite{mavali2024fake} adapts a simple defense method for CLIP-based AIGI detectors, by replacing the pretrained CLIP-ViT-L/14 vision backbone with a robust version obtained through unsupervised adversarial fine-tuning~\cite{schlarmann2024robustclipunsupervisedadversarial}. Specifically, they adopt pre-trained CLIP fine-tuned with PGD-$l_\infty$ with epsilons of $\varepsilon = 2/255$ and $\varepsilon = 4/255$, denoted as UnivFD-AF2 and UnivFD-AF4, respectively. Considering that \cite{mavali2024fake} is only available for UnivFD, we evaluate robustness using UnivFD detectors. As shown in \cref{tab:robustuni}, UnivFD-AF provides robustness only against C\&W, while it fails to defend against PGD and AutoAttack. In contrast, TRIM consistently demonstrates robustness across various attacks and datasets, outperforming Robust-UnivFD by a large margin. We speculate that UnivFD-AF, which directly utilizes an encoder adversarially fine-tuned on general tasks, cannot effectively learn the distinction between clean and adversarial samples specific to the AIGI detection task.

\begin{table}[tbp]
  \centering
  \caption{Clean accuracy(\%) and adversarial robustness(\%) evaluation compared with UnivFD-AF.}
  \resizebox{0.8\linewidth}{!}{
    \begin{tabular}{cccccc|ccccc}
    \hline
    \multirow{2}{*}{Method} & \multicolumn{5}{c|}{Dataset: ProGAN}   & \multicolumn{5}{c}{Dataset: GenImage} \\
\cline{2-11}          & Clean & PGD   & C\&W  & AA    & Ave   & Clean & PGD   & C\&W  & AA    & Ave \\
    \hline
    UnivFD-AF2 & 80.96 & 3.36  & 76.63 & 1.92  & 40.72  & 79.8  & 11.21 & 74.97 & 8.21  & 43.55  \\
    UnivFD-AF4 & 80.96 & 3.36  & 76.73 & 2.11  & 40.79  & 79.8  & 11.04 & \textbf{75.68} & 8.11  & 43.66  \\
    TRIM  & \textbf{98.18} & \textbf{100} & \textbf{93.88} & \textbf{87.62} & \textbf{94.92 } & \textbf{85.19} & \textbf{100} & 60.3  & \textbf{71.97} & \textbf{79.37 } \\
    \hline
    \end{tabular}%
    }
  \label{tab:robustuni}%
  \vspace{-0.8em}
\end{table}%

\begin{table}[!tbp]
  \centering
  \caption{Robustness evaluation against adaptive attacks.}
  \resizebox{0.9\linewidth}{!}{
    \begin{tabular}{cc|ccc|ccc|ccc}
    \hline
    \multirow{2}{*}{Dataset} & \multirow{2}{*}{Detector} & \multicolumn{3}{c|}{Original} & \multicolumn{3}{c|}{Entropy} & \multicolumn{3}{c}{TRIM} \\
\cline{3-11}          &       & Adapt1 & Adapt2 & Ave   & Adapt1 & Adapt2 & Ave   & Adapt1 & Adapt2 & Ave \\
    \hline
    \multirow{4}{*}{ProGAN} & CNNSpot & 0.00  & 0.90  & 0.45  & 34.80  & 4.03  & 19.42  & 92.07  & 90.83  & \textbf{91.45 } \\
          & UnivFD & 0.39  & 31.31  & 15.85  & 17.17  & 33.46  & 25.32  & 91.18  & 83.75  & \textbf{87.47 } \\
          & FreqNet & 0.09  & 7.17  & 3.63  & 31.40  & 17.80  & 24.60  & 74.87  & 69.34  & \textbf{72.11 } \\
          & NPR   & 0.30  & 3.12  & 1.71  & 28.08  & 25.72  & 26.90  & 94.78  & 85.78  & \textbf{90.28 } \\
    \hline
    \multirow{4}{*}{GenImage} & CNNSpot & 0.00  & 11.29  & 5.65  & 13.76  & 22.63  & 18.20  & 73.31  & 81.05  & \textbf{77.18 } \\
          & UnivFD & 0.93  & 7.76  & 4.35  & 17.93  & 10.02  & 13.98  & 57.80  & 57.75  & \textbf{57.78 } \\
          & FreqNet & 0.09  & 0.09  & 0.09  & 38.63  & 34.89  & 36.76  & 74.65  & 82.92  & \textbf{78.79 } \\
          & NPR   & 0.00  & 0.29  & 0.15  & 12.70  & 48.78  & 30.74  & 75.94  & 99.41  & \textbf{87.68 } \\
    \hline
    \end{tabular}
    }
  \label{tab:adaptive}
  \vspace{-0.8em}
\end{table}
\subsection{Robustness Evaluation Against Adaptive Attacks}
\label{sec:adaptive}
%In \cref{section_entropy} , we proposed an entropy-based defense that detects adversarial samples by identifying abnormal prediction entropy values. However, we observe that adaptive attacks can successfully bypass this defense by either constraining the perturbation magnitude or optimizing the logits within a subtle range. 

Because our defense mechanism relies on entropy bounds, we design two types of adaptive attacks that forces adversarial samples to exhibit entropy values similar to benign inputs by either constraining the perturbation magnitude or optimizing the logits within a subtle range. (1) \textbf{Adapt1}: PGD with $\epsilon=1/255$; and (2) \textbf{Adapt2}: C\&W attack with $\kappa=0$. Under these two attack settings, we conduct an ablation study to evaluate the performance of the standard detector, the naive entropy-based defense proposed in \cref{section_entropy}), and our proposed TRIM. As shown in Table~\ref{tab:adaptive}, Adapt1 and Adapt2 effectively bypass the original detector and the entropy-based defense. The original detector fails completely in most cases, with accuracy dropping close to zero. While the entropy-based method slightly improves robustness, its performance remains unstable and insufficient against adaptive attacks. In contrast, TRIM consistently achieves significantly higher accuracy across all models and datasets. For example, on CNNSpot with ProGAN, TRIM boosts the average accuracy from 0.45 (Original) and 19.42 (Entropy) to 91.45. These results demonstrate that TRIM is highly robust even under adaptive attacks designed to evade entropy-based defense.

%We design two types of adaptive attacks: (1) \textbf{Adapt1}: PGD with $\epsilon=1/255$; and (2) \textbf{Adapt2}: C\&W attack with $\kappa=0$. Under these two attack settings, we evaluate the performance of the original detector without defense, the entropy-based method, and our proposed TRIM. As shown in Table~\ref{tab:adaptive}, Adapt1 and Adapt2 effectively bypass the original detector and the entropy-based defense. The original detector fails completely in most cases, with accuracy dropping close to zero. While the entropy-based method slightly improves robustness, its performance remains unstable and insufficient against adaptive attacks. In contrast, TRIM consistently achieves significantly higher accuracy across all models and datasets. For example, on CNNSpot with ProGAN, TRIM boosts the average accuracy from 0.45 (Original) and 19.42 (Entropy) to 91.45. These results demonstrate that TRIM is highly robust even under adaptive attacks designed to evade entropy-based defense.

\subsection{Robustness Evaluation in Real-World Scenarios.}
\begin{table}[tbp]
% \vspace{-1em}
  \centering
  \caption{Clean and robust accuracy(\%) evaluation under the cross-generator image detection task for TRIM. The arrows indicate changes in clean accuracy compared to standard detectors.}
  \setlength{\tabcolsep}{5pt}
  \resizebox{1.0\linewidth}{!}{
    \begin{tabular}{cc|cccc|cccc|cccc}
    \hline
    \multirow{3}{*}{Training Set} & \multirow{3}{*}{Detector} & \multicolumn{12}{c}{Testing Set} \\
    \cline{3-14}
    & & \multicolumn{4}{c|}{ProGAN}   & \multicolumn{4}{c|}{GenImage} & \multicolumn{4}{c}{Stable Diffusion v1.4} \\
\cline{3-14}          &       & \multicolumn{2}{c}{Clean} & \multicolumn{1}{l}{PGD} & APGD    & \multicolumn{2}{c}{Clean} & PGD   & APGD    & \multicolumn{2}{c}{Clean} & PGD   & APGD \\
    \hline
    \multirow{4}[2]{*}{ProGAN} & CNNSpot & 98.9 & $\downarrow{0.99}$  & 100   & 99.74 & 54.81&$\downarrow{0.84}$  & 100   & 71.75 & 56.4&$\uparrow{4.67}$ & 100   & 66.87 \\
          & UnivFD & 98.18&$\downarrow{0.58}$  & 100   & 96.62 & 66.4&$\uparrow{1.92}$ & 99.4  & 68.85 & 54.66&$\uparrow{0.05}$ & 99.65 & 56.55 \\
          & FreqNet & 93.47&$\downarrow{2.31}$  & 95.05 & 93.37 & 64.08&$\downarrow{1.69}$  & 95.47 & 78.83 & 60.36&$\uparrow{6.79}$ & 96.36 & 70.59 \\
          & NPR   & 92.22&$\downarrow{3.46}$  & 99.98 & 99.94 & 65.2&$\downarrow{3.83}$  & 99.61 & 95.37 & 66.47&$\downarrow{5.05}$  & 99.39 & 95.53 \\
    \hline
    \end{tabular}
    }
  \label{tab:cross}
  \vspace{-0.8em}
\end{table}

\textbf{Cross-Generator Generalization Analysis.} In practical scenarios, AIGI detectors are inevitably exposed to images generated by a wide range of generative models. Since the source generator of a given image is typically unknown at inference time, it is essential to assess whether the proposed defense strategy can generalize to unseen generators. To this end, we pre-train detectors on one dataset and evaluate their accuracy and robustness on others. For TRIM, we fix the detection threshold on the ProGAN dataset and do not adjust it for other generators. We observe that robustness is positively correlated with generalization on clean samples. As shown in Table~\ref{tab:cross}, training and testing on the same dataset yield high robustness and accuracy. In comparison, detectors that generalize poorly to unseen generators on clean samples also show degraded robustness. However, TRIM's robustness generalization surprisingly surpasses the clean generalization of standard detectors. For instance, when trained on ProGAN and tested on GenImage and SD v1.4, the clean accuracy of standard detectors drops to 61.08\%(decreasing 34.61\% compared to the results in ProGAN dataset), while TRIM maintains 87.13\% robustness (only 10.95\% drop). These results highlight TRIM's strong generalization in robustness across unseen generators.

\begin{figure}[tbp]
    \centering
    \includegraphics[width=\linewidth]{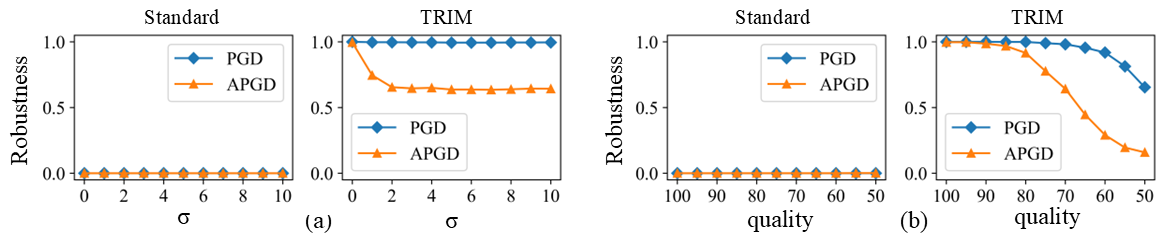}
    % \vspace{-0.5em}
    \caption{(a) Gaussian blur; (b) JPEG compression. Each subfigure compares the standard detector without defense and TRIM, both trained on ProGAN using CNNSpot.}
    \label{fig:blurandcompress}
    \vspace{-1em}
\end{figure}

\textbf{Robustness under Image Quality Degradations.} In real-world scenarios, images may undergo various forms of quality degradation before being processed by detection systems. For instance, platforms such as image-sharing websites or content moderation systems often introduce automatic compression or resizing, unintentionally altering the original inputs. Therefore, we conduct experiments on adversarial samples under common image degradation conditions, including JPEG compression and Gaussian blur, following the settings in GenImage\cite{zhu2023genimage}. Specifically, we apply JPEG compression with quality levels ranging from 100 to 50, and Gaussian blur with standard deviation $\sigma$ varying from 1 to 10. The results are presented in  \cref{fig:blurandcompress}. Although robustness declines to some extent as degradation severity increases, TRIM still largely improve the robustness compared to standard detectors.  

%Nevertheless, TRIM still largely improve the robustness compared to standard detectors under the degradation setting.

%As illustrated in \cref{fig:blurandcompress}, robustness declines to some extent as degradation severity increases. Notably, quality levels below 60 exhibit poor image quality, making visual content difficult to discern (see Appendix C for visual results). Consequently, the significant drop in robustness is understandable.

%Compared with the original detector without defense, TRIM consistently shows robustness under most image degradation conditions.
% Notably, adversarial samples generated by PGD and AutoAttack remain highly effective on the standard, undefended detector, consistently succeeding even under substantial input distortions. In contrast, TRIM still provides strong defense against such resilient attacks under these distortion scenarios.
%attackers might deliberately apply compression or blur to conceal adversarial perturbations. Meanwhile,

% \begin{figure}[htbp]
%     \centering
%     \begin{subfigure}[b]{\textwidth}
%         \includegraphics[width=\linewidth]{pictures/compress.png}
%         \caption{compression}
%         \label{fig:sub1}
%     \end{subfigure}
    
%     % \vspace{-0.5em}

%     \begin{subfigure}[b]{\textwidth}
%         \includegraphics[width=\linewidth]{pictures/blur.png}
%         \caption{blur}
%         \label{fig:sub2}
%     \end{subfigure}
    
%     \caption{compression and blur}
%     \label{fig:overall}
% \end{figure}

\vspace{-0.8em}
\section{Conclusion and Outlook}
We propose TRIM, the first training-free adversarial defense for AIGI detection. Its model-agnostic and training-free design, as well as robustness against various attack even under cross-generator and image degradation scenarios make TRIM a practical solution for real-world applications. It fills the gap of lacking defenses for AIGI detection and calls for attention to adversarial robustness in the detection of AI-generated content (AIGC). One limitation is that the thresholds in TRIM need to be manually configured. However, this is lightweight as the distinct feature shift between clean and adversarial data makes threshold selection straightforward. Moreover, a single set of thresholds can generalize well across different datasets. TRIM has the potential to be extended to other types of generated content, as they are often produced by similar generator architectures (e.g., GANs or Diffusion) and may therefore share common artifact clues. In future work, we will explore methods to enhance the adversarial robustness of AIGC detection across multiple modalities and data types, including AI-generated video and audio.

% TRIM has the potential to be extended to other synthetic media formats, such as AI-generated video and audio. In future work, we will try to extend TRIM to 

% aim to continue exploring ways to enhance the adversarial robustness of AIGC detection across multiple modalities and data types.
%We propose TRIM, the first training-free adversarial defense for AIGI detection that employs information-theoretic quantification to approximate inaccessible real-world adversarial feature shifts, enabling conversion of standard detectors into robust ones without retraining. Extensive experiments demonstrate that TRIM effectively defends against a wide range of attacks while maintaining high accuracy on clean inputs, outperforming state-of-the-art defense methods, including both AT and test-time defenses. TRIM offers a lightweight, model-agnostic solution for improving adversarial robustness in AIGI detection.

\appendix
\section{Related Work}
\label{appendix:related work}
\subsection{AI-generated Image Detection}
The demand for distinguishing between real and fake images has been steadily increasing. Recent research has focused on diverse discriminative features from different perspectives. Early works \cite{wang2020cnn,liu2020global,zhang2019detecting,frank2020leveraging} adopt a data-driven approach to end-to-end learning of artifacts induced by generator upsampling operations, capturing patterns in spatial or frequency domains. In order to further improve the generalization capability of detection methods on images produced by various generative models UnivFD and FatFormer\cite{liu2023forgeryawareadaptivetransformergeneralizable,liu2024forgery} extract semantic features using pre-trained CLIP; Dire, AEROBLADE \cite{zhang2023diffusion,ricker2024aeroblade} introduce reconstruction error from the pretrained diffusion model as discriminative cues, while some methods\cite{zhong2023patchcraft,liu2022detecting,tan2023learning,tan2024rethinking} extract universal artifact representations by leveraging pixel-level fingerprints. These approach the real-versus-fake image classification problem from various angles, enriching the research paradigm of image authenticity detection.
\subsection{Adversarial Robustness of AIGI detectors}
\cite{de2024exploring} systematically evaluates the adversarial robustness of AIGI detectors based on CNN and CLIP under different attack(such as PGD\cite{madry2019deeplearningmodelsresistant}, and UA\cite{moosavi2017universal}), and finds that both models are vulnerable to white-box attacks. \cite{mavali2024fake} further shows that even if the attacker cannot access the target model and the adversarial samples are post-processed by compression, they can still effectively attack SOTA detectors in real scenarios, significantly weakening their detection capabilities and increasing the risk of misjudgment. In the field of forgery detection, certain adversarial attack methods against AIGI detection pose significant challenges to AI-generated image (AIGI) detectors by targeting their reliance on frequency domain features. These attacks aim to reduce the spectral or statistical frequency differences between real and generated images\cite{dong2022think}\cite{hou2023evading}\cite{jia2022exploring}\cite{zhou2024stealthdiffusion}, thereby misleading detectors that depend on such cues. Among them, Stealthdiffusion\cite{zhou2024stealthdiffusion} optimizes adversarial perturbations in both spatial and frequency domains, posing a strong threat to detectors relying on cues of either domain, and stands out as a state-of-the-art attack against AIGI detectors.

\subsection{Adversarial Defense}
% Adversarial Training (AT)\cite{madry2019deeplearningmodelsresistant}, proposed by Goodfellow, is a classic and effective defense that enhances robustness by training on adversarial samples to learn attack patterns and refine decision boundaries. Building on this framework, \cite{madry2019deeplearningmodelsresistant} formalized AT as a min-max optimization framework (PGD-AT), generating adversarial samples via PGD in the inner maximization step and updating the model by minimizing losses on these samples. \cite{zhang2019theoreticallyprincipledtradeoffrobustness} introduced TRADES, which theoretically decomposes the objective into natural classification error and a KL divergence term penalizing discrepancies between natural and adversarial predictions.

To address the vulnerability of deep models to adversarial examples, adversarial training has emerged as a mainstream defense strategy. PGD-AT \cite{madry2019deeplearningmodelsresistant} formulates a min-max optimization framework by training models on adversarial samples crafted via PGD, significantly improving robustness under white-box attacks. However, this approach often incurs a considerable drop in clean accuracy. To better balance robustness and generalization, TRADES \cite{zhang2019theoreticallyprincipledtradeoffrobustness} introduces a loss function that decomposes adversarial risk into natural loss and KL divergence between predictions on clean and adversarial samples. To balance robustness and accuracy, as well as to improve robust generalization, RAT\cite{jin2023randomizedadversarialtrainingtaylor} adds random noise to deterministic weights, leading to smoother weight updates and enabling the model to converge to flatter minima.

Beyond training-phase defenses, test-time defense mechanisms aim to mitigate adversarial threats during inference, often through detection or purification. Adversarial detection identifies adversarial samples by comparing their intrinsic differences from clean inputs. As a classic paradigm, \cite{Xu_2018} detects adversarial samples by comparing model predictions on original inputs versus compressed ones (e.g., color depth reduction and Gaussian blur), while other works\cite{abusnaina2021adversarial}\cite{ma2018characterizing}\cite{zhang2024detecting} exploit image reconstruction quality and intrinsic dimensionality to distinguish adversarial inputs. Adversarial purification mitigates the impact of adversarial perturbations by transforming inputs at inference time. Among existing approaches, diffusion models have become a mainstream solution for purification, as demonstrated in \cite{dolatabadi2024devil, may2023salient, wang2022guided}. These methods leverage the generative capacity of diffusion models to reconstruct clean versions of adversarial inputs. TPAP\cite{tang2024robust} recently proposed a novel approach that employs overfitted FGSM adversarially trained model to reverse-process test images using FGSM attacks, effectively purifying unknown perturbations.

\section{Proof}
\label{appendix:proof}
\subsection{Proof for Proposition 1.}
In this section, we provide the proof of Proposition 1. in Section 2.

\textbf{Proposition 1.} Let $ Z, \tilde{Z}, \Delta Z, Y $ denote four random variables, where $\tilde{Z}=Z+\Delta Z$. Then the following approximate relationship holds:
\begin{equation}
    I(\tilde{Z};Y)\approx I(Z;Y)+I(\Delta Z;Y|Z).
\end{equation}

\textbf{Proof. } 
%The following proof mainly refers to \cite{zhou2022improvingadversarialrobustnessmutual}, 

According to the relationship between information entropy and mutual information, we first provide the following lemma.

\textbf{Lemma 1:}
\begin{equation}
\left\{
\begin{aligned}
I(\tilde{Z}; Y) &= H(Y) - H(Y|\tilde{Z}) \\
I(Z; Y) &= H(Y) - H(Y|Z) \\
I(\Delta Z; Y) &= H(Y) - H(Y|\Delta Z)
\end{aligned}
\right.
\end{equation}

where $ H(\cdot) $ denotes the information entropy, and $ H(\cdot | \cdot) $ denotes the conditional information entropy.

According to Lemma 1, we have:
\begin{equation}
  \begin{aligned}
I(Z; Y) + I(\Delta Z; Y) &= H(Y) - H(Y|Z) + H(Y) - H(Y|\Delta Z) \\
&= 2H(Y) - [H(Y|Z) + H(Y|\Delta Z)].
\label{eq:varia1}
\end{aligned}  
\end{equation}

According to the theorem of conditional mutual information in probability theory, we have:
\begin{equation}
 \begin{aligned}
H(Y|Z) + H(Y|\Delta Z) &= [H(Y|Z, \Delta Z) + I(Y; \Delta Z | Z)] + [H(Y|\Delta Z, Z) + I(Y; Z | \Delta Z)] \\
&= 2H(Y|Z, \Delta Z) + I(Y; \Delta Z | Z) + I(Y; Z | \Delta Z) \\
&=[H(Y|Z,\Delta Z)+I(Y;\Delta Z|Z)+I(Y;Z|\Delta Z)+I(Z;\Delta Z;Y)]+H(Y|Z,\Delta Z)-I(Z;\Delta Z;Y) \\
&= H(Y) + H(Y|Z, \Delta Z) - I(Z; \Delta Z; Y).
\label{eq:varia2}
\end{aligned}   
\end{equation}

Combining \cref{eq:varia1} and \cref{eq:varia2}, we have:
\begin{equation}
 \begin{aligned}
I(Z; Y) + I(\Delta Z; Y) &= 2H(Y) - [H(Y|Z) + H(Y|\Delta Z)] \\
&= H(Y) - H(Y|Z, \Delta Z) + I(Z; \Delta Z;Y) \\
&= I(\tilde{Z}; Y) + H(Y|\tilde{Z}) - H(Y|Z, \Delta Z) + I(Z; \Delta Z;Y).
\end{aligned}   
\end{equation}

Finally, we have:
\begin{equation}
I(\tilde{Z}; Y) = I(Z; Y) + I(\Delta Z; Y) - I(Z; \Delta Z;Y) + H(Y|Z, \Delta Z) - H(Y|\tilde{Z}).
\label{eq:final}
\end{equation}

Since $Z + \Delta Z$ can completely reconstruct $\tilde{Z}$, i.e., $\tilde{Z}$ is fully determined by $Z$ and $\Delta Z$, and the combination $Z + \Delta Z$ contains all the information of both $Z$ and $\Delta Z$~\cite{zhou2022improvingadversarialrobustnessmutual}, we assume the conditional entropy $H(Y | Z, \Delta Z)$ and $H(Y | \tilde{Z})$ are approximately equal. Consequently, \cref{eq:final}can be can be simplified to
\begin{equation}
\begin{aligned}
    I(\tilde{Z};Y)&\approx I(Z;Y)+I(\Delta Z;Y)-I(Z;\Delta Z;Y).
    \label{eq:AT_result}
\end{aligned}    
\end{equation}

According to the expansion formula of the multivariate mutual information, we have:
\begin{equation}
    I(Z; \Delta Z; Y)=I(\Delta Z; Y) - I(\Delta Z; Y | Z) .
\label{eq:multivariate}
\end{equation}

Combining \cref{eq:AT_result} and \cref{eq:multivariate}, we have:

\begin{equation}
    I(\tilde{Z};Y)\approx I(Z;Y)+I(\Delta Z; Y | Z)
    \label{eq:prop1}
\end{equation}
which completes the proof.

\subsection{Proof for Proposition 2.}
In this section, we provide the proof of Proposition 2. in Section 3.1.

\textbf{Proposition 2.}
Let $\tilde{y}$ denote the softmax output of an AIGI detector and $\tilde{Z}$ the adversarial feature. 
\begin{itemize}
\item For successful adversarial samples(misclassified), the model's prediction is mainly influenced by the feature shift $\Delta Z$  caused by adversarial perturbation, i.e.,
 $I(\tilde{Z}; \tilde{y}) \approx I(\Delta Z; \tilde{y})$.
 
Using $I(A; B) = H(B) - H(B|A)$, we have:
\begin{equation}
H(\tilde{y} | \tilde{Z}) \approx H(\tilde{y} | \Delta Z) .
\label{eq:entropy_success}
\end{equation}
\item For unsuccessful adversarial samples(correctly classified), the prediction remains largely governed by the original clean feature $Z$ , i.e., $I(\tilde{Z}; \tilde{y}) \approx I(Z; \tilde{y})$, which implies:
\begin{equation}
H(\tilde{y} | \tilde{Z}) \approx H(\tilde{y} | Z ).
\label{eq:entropy_fail}
\end{equation}
\end{itemize}
% The prediction entropy $H(\tilde{y} | \tilde{Z})$ for a single sample is computed as:
% \begin{equation}
% H(\tilde{y} | \tilde{Z}) = -\sum_{y_i \in \mathcal{Y}} p(y_i | \tilde{Z}) \log p(y_i  | \tilde{Z}),
% \end{equation}
% where $\mathcal{Y}$ denotes the set of all possible labels in the dataset and the probabilities are obtained from the model’s softmax output.

\textbf{Proof. }According to \cref{eq:AT_result}, we have the following corollary:

\textbf{Corollary 1:}
\begin{equation}
    I(\tilde{Z};\tilde y)\approx I(Z;\tilde y)+I(\Delta Z;\tilde y)-I(Z;\Delta Z;\tilde y).
\end{equation}

Although $I(\tilde{Z}; \tilde{y})$ is jointly influenced by $I(Z; \tilde{y})$, $I(\Delta Z; \tilde{y})$, and $I(Z; \Delta Z; \tilde{y})$, in fact, only one of these terms typically plays a dominant role.
As noted by~\cite{zhou2022improvingadversarialrobustnessmutual}, since the predictions are zero sum (either correct or incorrect) and the effects of natural and adversarial patterns on the output are mutually exclusive, $H(\tilde y)$ has a large overlap with $H(X)$ and a small overlap with $H(\Delta X)$ when adversarial examples are not successfully misclassified by the detector. Conversely, it has a large overlap with $H(\Delta X)$ and a small overlap with $H(X)$ when adversarial examples are successfully misclassified by the detector. Here, $X$ denotes the clean input, and $\Delta X$ represents the perturbation added to $X$ to obtain the adversarial example $\tilde{X}$, i.e., $\tilde{X} - X$.

\begin{figure}[htbp]
    \centering
    \includegraphics[width=\linewidth]{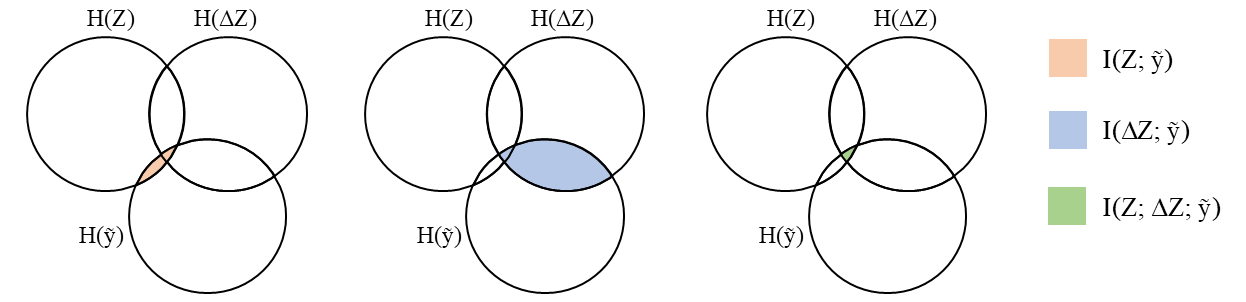}
    \caption{The visualization of $I(Z;\tilde y)$, $I(\Delta Z;\tilde y)$ and $I(Z; \Delta Z;\tilde y)$ of successful adversarial samples.}
    \label{fig:MI_success}
   \vspace{-0.8em}
\end{figure}

\begin{figure}[htbp]
    \centering
    \includegraphics[width=1\linewidth]{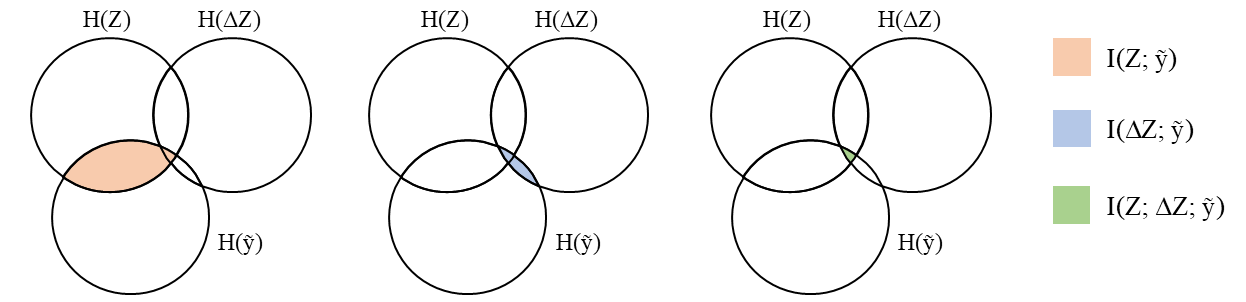}
    \caption{The visualization of $I(Z;\tilde y)$, $I(\Delta Z;\tilde y)$ and $I(Z; \Delta Z;\tilde y)$ of unsuccessful adversarial samples.}
    \label{fig:MI_unsuccess}
    \vspace{-0.8em}
\end{figure}

Since the model's output is directly derived from the compressed representation $\tilde{Z}$ of $\tilde{X}$, and these features retain all relevant information about the final prediction, i.e., $p(\tilde{y} |\tilde{X}) = p(\tilde{y} | \tilde{Z})$ and $H(\tilde{y} | \tilde{X}) = H(\tilde{y} | \tilde{Z})$
, we extend this assumption to the feature space. Specifically, when the attack succeeds, $H(\tilde{y})$ has a large overlap with $H(\Delta Z)$ and a small overlap with $H(Z)$ (\cref{fig:MI_success}); when the attack fails, $H(\tilde{y})$ has a large overlap with $H(Z)$ and a small overlap with $H(\Delta Z)$ (\cref{fig:MI_unsuccess}). Notably, in both cases, the mutual information $I(Z;\Delta{Z};\tilde{y})$ is small and can therefore be neglected.

Through the above analysis, we have:
\begin{equation}
I(\tilde{Z}; \tilde{y}) \approx
\begin{cases}
I(\Delta Z; \tilde{y}) & \text{(attack succeeds)} \\
I(Z; \tilde{y}) & \text{(attack fails)}
\end{cases}
\end{equation}

% Given that I(A|B)=H(B)-H(B|A), we have:
% \begin{equation}
% H(\tilde{y}|\tilde{Z}) \approx
% \begin{cases}
% H(\tilde{y}|\Delta Z) & \text{(attack succeeds)} \\
% H(\tilde{y}|Z) & \text{(attack fails)}
% \end{cases}
% \end{equation}

% In computing $H(\tilde{y} | \tilde{Z})$, we use the softmax output as the probability distribution. First, the softmax output of a classifier can be interpreted as a conditional probability distribution $p(y | X)$ over the discrete label set~\cite{smith2018understandingmeasuresuncertaintyadversarial}, where $X$ is the input to the classifier and $y$ represents a possible class label. Second, since the feature extraction from $X$ to the representation $Z$ is deterministic and the prediction $y$ is directly obtained from $Z$, the representation $Z$ preserves all information relevant for the prediction. Therefore, the conditional distributions satisfy $p(y | Z) = p(y | X)$, allowing us to use the softmax output based on $\tilde X$ to estimate $p(\tilde{y} | \tilde{Z})$.

% Combining these points, we conclude that the model’s softmax output can be directly employed to estimate $p(\tilde{y} | \tilde{Z})$, and thus the conditional entropy can be computed as:
% \begin{equation}
% H(\tilde{y} | \tilde{Z}) = -\sum_{y_i \in \mathcal{Y}} p(y_i | \tilde{Z}) \log p(y_i | \tilde{Z})=-\sum_{y_i \in \mathcal{Y}} p(y_i | \tilde{X}) \log p(y_i | \tilde{X}),
% \end{equation}
% where $\mathcal{Y}$ denotes the set of all possible labels.
which completes the proof.

\subsection{Proof for Proposition 3.}
In this section, we provide the proof of Proposition 3. in Section 3.2.

\textbf{Proposition 3.} Let $Z^b$ and $Z^a$ denote the feature before and after denoising, respectively, and let $y^b$ and $y^a$ represent the corresponding softmax outputs. Then, the following inequality holds:
\begin{equation} 
I(Z^b - Z^a; y^b | Z^a)  \approx H(y^b | Z^a) - H(y^b | Z^b) \  \leq D_{\text{KL}}(y^b | Z^b, y^a | Z^a),  
\end{equation}
where $D_{\text{KL}}(\cdot, \cdot)$ denotes the KL divergence between the detector outputs before and after adversarial denoising. The inequality becomes an equality if and only if $Z^a = Z^b$.

\textbf{Proof. }
According to \cref{eq:prop1}, we have the following corollary:

\textbf{Corollary 2. }
\begin{equation} 
I(Z^b - Z^a; y^b | Z^a) \approx I(Z^b;y^b)-I(Z^a;y^b) = H(y^b | Z^a) - H(y^b | Z^b)
\label{eq:coro2}
\end{equation}

Since $H(y^b | Z^a)$ cannot be calculated directly, we use cross entropy $H_{cross}(y^b|Z^{b},y^a|Z^{a})$ as an upper bound to estimate this $H(y^b | Z^a)$:

\begin{equation}
\begin{aligned}
H_{cross}(y^b|Z^{b},y^a|Z^{a})&=E_{z^{b},z^{a}\sim p(Z^{b},Z^{a})}(-\sum_{y \in \mathcal{Y}}p(y^b|z^b)log(p(y^a|z^a)))\\
&=E_{z^a \sim p(Z^a)}(-\sum_{z^b}p(z^b|z^a)\sum_{y \in \mathcal{Y}}p(y^b|z^b)log(p(y^a|z^a)))\\
&=E_{z^a \sim p(Z^a)}(-\sum_{y \in \mathcal{Y}}\sum_{z^b}p(z^b|z^a)p(y^b|z^b)log(p(y^a|z^a)))
\label{eq:kl_1}
\end{aligned}
\end{equation}

Considering that $y^b$ is fully determined by $Z^b$, i.e., given $Z^b$, $y^b$ is conditionally independent of $Z^a$, we have: $p(y^b \mid Z^b) = p(y^b \mid Z^b, Z^a)$, therefore:

\begin{equation}
\begin{aligned}
H_{cross}(y^b|Z^{b},y^a|Z^{a})
&=E_{z^a \sim p(Z^a)}(-\sum_{y \in \mathcal{Y}}\sum_{z^b}p(z^b|z^a)p(y^b|z^b)log(p(y^a|z^a)))\\
&=E_{z^a \sim p(Z^a)}(-\sum_{y \in \mathcal{Y}}\sum_{z^b}p(z^b|z^a)p(y^b|z^b,z^a)log(p(y^a|z^a)))\\
&=E_{z^a \sim p(Z^a)}(-\sum_{y \in \mathcal{Y}}p(y^b|z^a)log(p(y^a|z^a)))\\
&=E_{z^a \sim p(Z^a)}(-\sum_{y \in \mathcal{Y}}p(y^b|z^a)log(p(y^b|z^a))+\sum_{y \in \mathcal{Y}}p(y^b|z^a)log\frac {p(y^b|z^a)}{p(y^a|z^a)})\\
&=H(y^b|Z^a)+KL(p(y^b|Z^a),p(y^a|Z^a))\\
&\geq H(y^b|Z^a)
\label{eq:kl_2}
\end{aligned}
\end{equation}
Only when $ p(y^b|Z^a) = p(y^a|Z^a) $, the KL divergence becomes zero, and the inequality turns into an equality. In other words, the equality $ H_{\text{cross}}(y^b|Z^b, y^a|Z^a) = H(y^b|Z^a) $ holds if and only if the preprocessed feature $ Z^a $ retains all the discriminative information originally present in $ Z^b $, i.e., when $ Z^a $ and $ Z^b $ are equivalent in terms of predictive power.

Combining \cref{eq:coro2}, \cref{eq:kl_1}, and \cref{eq:kl_2}, we have the following:
\begin{equation}
    I(Z^b - Z^a; y^b | Z^a)  \approx H(y^b | Z^a) - H(y^b | Z^b)\leq H_{cross}(y^b|Z^{b},y^a|Z^{a})-H(y^b | Z^b)
\label{eq:halb}
\end{equation}

Considering that $\mathrm{D_{KL}}(P \parallel Q) = H_{\text{cross}}(P, Q) - H(P)$, we have:

\begin{equation}
     H_{cross}(y^b|Z^{b},y^a|Z^{a})-H(y^b | Z^b)=\mathrm{D_{KL}}(p(y^b | Z^b), p(y^a | Z^a))
\label{eq:cross_kl}
\end{equation}

Combining \cref{eq:halb} and \cref{eq:cross_kl}, we have:
\begin{equation} 
I(Z^b - Z^a; y^b | Z^a)  \approx H(y^b | Z^a) - H(y^b | Z^b) \  \leq D_{\text{KL}}(y^b | Z^b, y^a | Z^a),  
\end{equation}

which completes the proof.

% However, in practice, the preprocessing step inevitably leads to a loss of label-relevant information. As a result, $ p(y^b|Z^a) \neq p(y^a|Z^a) $, which causes the KL divergence term to be positive. This implies that the cross-entropy $ H_{\text{cross}} $ is strictly greater than the conditional entropy $ H(y^b|Z^a) $. Therefore, using $ Z^a $ to predict $ y^b $ introduces additional uncertainty, effectively inflating the original entropy. 

% \subsection{for A Single Smaple}
% \begin{equation}
%     \begin{aligned}
% H_{\mathrm{cross}} & =-\sum_yp(y^b|Z^b)\log p(y^a|Z^a) \\
%  & =\underbrace{-\sum_yp(y^b|Z^b)\log p(y^b|Z^a)}_{1}+\underbrace{\sum_yp(y^b|Z^b)\log\frac{p(y^b|Z^a)}{p(y^a|Z^a)}}_{2}
% \end{aligned}
% \end{equation}

% \begin{equation}
%     \begin{aligned}
% 1 & =-\sum_y\left(\sum_{z^b}p(z^b|Z^a)p(y^b|z^b)\right)\log p(y^a|Z^a) \\
%  & =-\sum_{z^b}p(z^b|Z^a)\sum_yp(y^b|z^b)\log p(y^a|Z^a) \\
%  & =\sum_{z^b}p(z^b|Z^a)\left(-\sum_yp(y^b|z^b)\log p(y^a|Z^a)\right)\\
%  &\geq \sum_{z^b}p(z^b|Z^a)\left(-\sum_yp(y^b|z^b)\log p(y^a|z^b)\right)
% \end{aligned}
% \end{equation}

\section{Additional Experiment Details}
\label{appendix:additional Experiment Details}
\subsection{Computational Resources}
Our experiments involve training AIGI detectors, generating adversarial examples, and evaluating both clean accuracy and adversarial robustness. All experiments are conducted on two NVIDIA GeForce RTX 3090 GPUs (each with 24GB of VRAM). 

\subsection{Estimating $I(Z; Y)$ , $I(\tilde Z; Y)$ and $I(\Delta Z; Y | Z)$}
In this section, we detail the estimation procedures used to approximate mutual information values $I(Z; Y)$ , $I(\tilde Z; Y)$ and $I(\Delta Z; Y | Z)$ throughout adversarial training.

To evaluate the discriminative power of the learned representation $Z$ with respect to the true label $Y$, we consider the mutual information $I(Z; Y)$, defined as:
\begin{equation}
    I(Z; Y) = H(Y) - H(Y|Z),
\end{equation}
where $H(Y)$ is the marginal entropy of the labels, which is a fixed constant determined by the empirical label distribution, and $H(Y|Z)$ is the conditional entropy of $Y$ given the representation $Z$.

Since $H(Y|Z)$ is generally intractable to compute exactly, we approximate it using the cross-entropy~\cite{achille2018emergence,achille2018information,amjad2019learning} between the true labels and the model's predicted probabilities. Specifically, We approximate the conditional entropy as:
\begin{equation}
\begin{aligned}
    H(Y|Z) &\approx \mathbb{E}_{(x, y) \sim \mathcal{D}} [-\log p(y|x)] \\
    &= \text{CrossEntropy}(Y, \hat{Y}),
\end{aligned}
\end{equation}
where $(x, y) \sim \mathcal{D}$ denotes samples drawn from the training dataset, and $p(y|x)$ is the model’s predicted probability for the true label $y$.

% Since $z$ is deterministically computed from the input $x$, and the classifier operates directly on $z$, the predictive distribution is conditionally equivalent: $p(y|z) = p(y|x)$.
% This allows us to rewrite the expectation over $p(y|z)$ as an expectation over the model's output on clean inputs:
% \begin{equation}
%     H(Y|Z) \approx \mathbb{E}_{(x, y) \sim \mathcal{D}} [-\log p(y|x)] = \text{CrossEntropy}(Y, \hat{Y}),
% \end{equation}
% where $\hat{Y}$ denotes the model's predicted probability distribution over labels for input $X$.

Combining the above, we obtain the following approximation for the mutual information:
\begin{equation}
    I(Z; Y) \approx H(Y) - \text{CrossEntropy}(Y, \hat{Y}).
\end{equation}

% This approximation suggests that the mutual information increases as the model becomes more predictive (i.e., as the cross-entropy loss decreases), indicating that the learned representation $Z$ contains more information relevant to the label $Y$.

% Given that the marginal entropy $H(Y)$ is a constant determined by the empirical label distribution, we approximate the mutual information $I(Z; Y)$ as:
% \begin{equation}
%     I(Z; Y) = H(Y) - H(Y|Z) \approx H(Y) - \mathbb{E}_{(z, y) \sim \mathcal{D}} [-\log p(y|z)],
% \end{equation}
% where $p(y|z)$ is approximated by the model's predicted probability (i.e., softmax output) and the expectation is approximated by averaging over the training set. This simplifies to:
% \begin{equation}
%     I(Z; Y) \approx H(Y) - \text{CrossEntropy}(Y, \hat{Y}),
% \end{equation}
% where $\hat{Y}$ is the model prediction on clean inputs.

Next, We use the decomposition:
\begin{equation}
    I(\Delta Z; Y | Z) = I(\tilde{Z}; Y) - I(Z; Y),
\end{equation}
where $\tilde{Z} = Z + \Delta Z$ is the feature representation of adversarial inputs. $I(\tilde{Z}; Y)$ is estimated in the same way as $I(Z; Y)$, but using adversarial inputs:
\begin{equation}
    I(\tilde{Z}; Y) \approx H(Y) - \text{CrossEntropy}(Y, \hat{Y}_{\text{adv}}),
\end{equation}
where $\hat{Y}_{\text{adv}}$ is the model prediction on adversarial samples.

Therefore, we approximate $I(\Delta Z; Y | Z)$ as:
\begin{equation}
    I(\Delta Z; Y | Z) \approx \text{CrossEntropy}(Y, \hat{Y}) - \text{CrossEntropy}(Y, \hat{Y}_{\text{adv}}).
\end{equation}

These estimates allow us to visualize and compare the dynamic behaviors of clean feature learning and adversarial robustness during training.

\subsection{Additional Experiment Results of AT}
\begin{figure}[htbp]
    \centering
    \includegraphics[width=\linewidth]{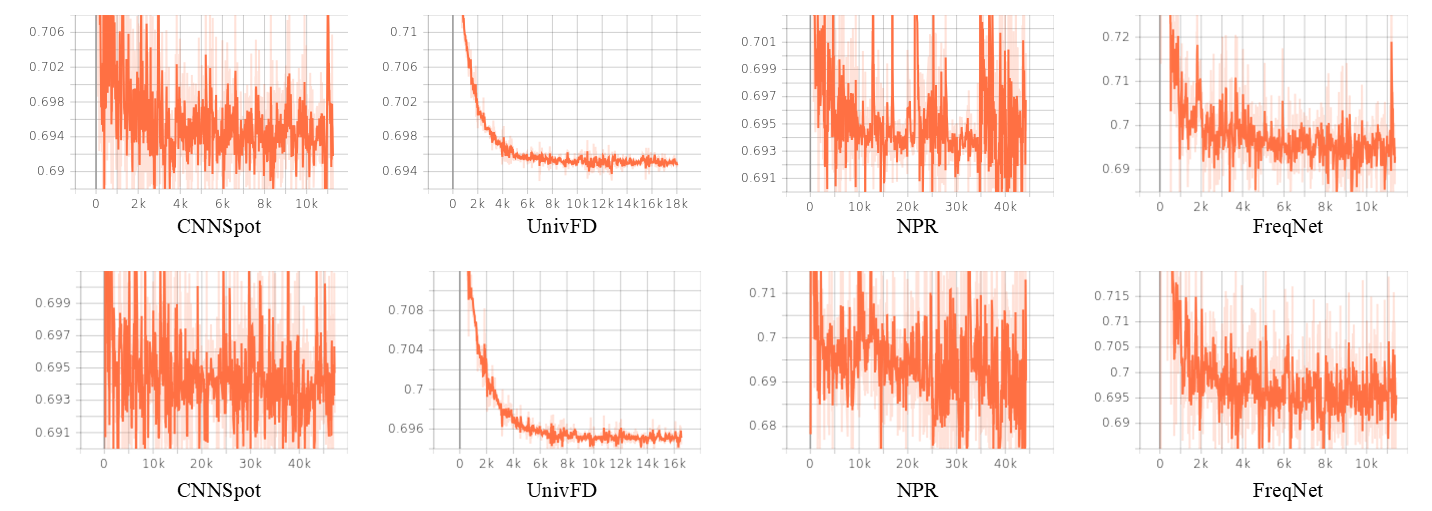}
    \caption{Convergence of loss of PGD-AT on ProGAN dataset (first row) and GenImage (second row)}
    \label{fig:pgdat_loss}
% \vspace{-1em}
\end{figure}

\begin{figure}[htbp]
    \centering
    \includegraphics[width=\linewidth]{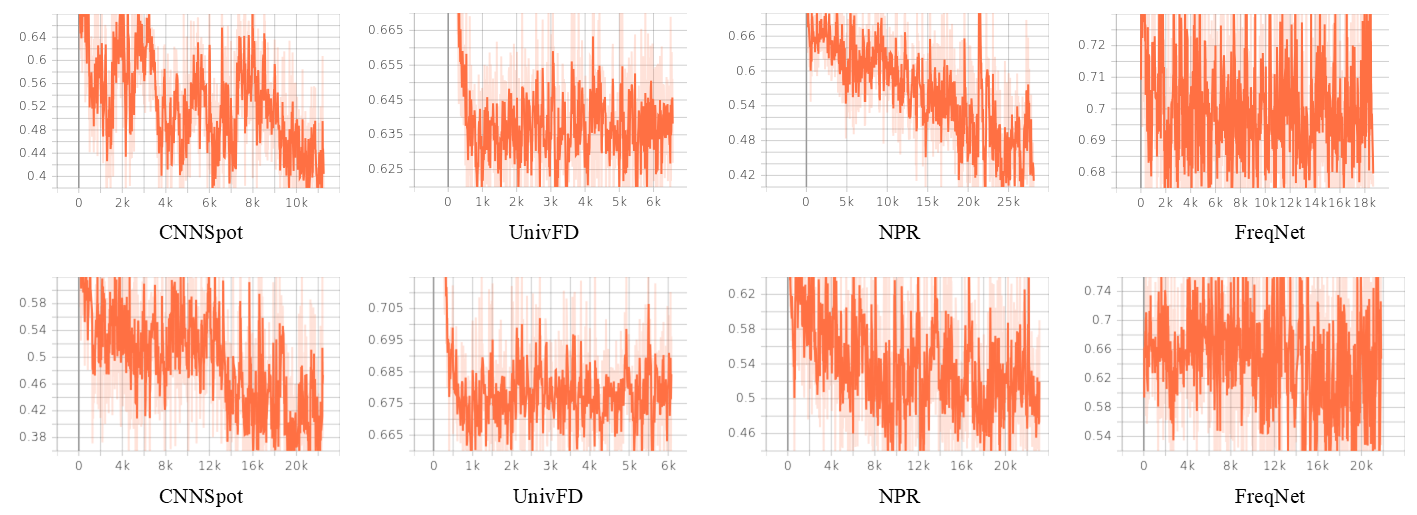}
    \caption{Convergence of loss of TRADES on ProGAN dataset (first row) and GenImage (second row)}
    \label{fig:trades_loss}
    % \vspace{-1em}
\end{figure}

In this section, we present additional experimental results related to adversarial training as discussed in Section 2. As shown in \cref{fig:pgdat_loss} and \cref{fig:trades_loss}, we observe that both PGD-AT and TRADES fail to achieve effective training on ProGAN and GenImage datasets when using the default perturbation strength of $8/255$. For PGD-AT, the loss fails to converge and fluctuates around 0.69 (approximately $\ln 2$), indicating that the model is essentially making random predictions. In the case of TRADES, although the loss decreases slightly in some instances, there is no improvement in either robustness or accuracy on the validation set, ultimately leading to early termination of training due to early stopping.
 To further investigate whether reducing the perturbation magnitude can mitigate this issue, we conduct adversarial training on CNNSpot with ProGAN dataset using different perturbation strengths. The results are summarized in \cref{tab:at}. As shown, even with reduced perturbation magnitudes, adversarial training still fails to converge successfully.
 
\begin{table}[tbp]
  \centering
  \caption{Clean accuracy and adversarial robustness(\%) under adversarial training of different intensities}
  \resizebox{1.0\linewidth}{!}{
    \begin{tabular}{cccccccccccc}
    \toprule
          &       &       & \multicolumn{3}{c}{Clean Samples} & \multicolumn{2}{c}{PGD;eps=8/255} & \multicolumn{2}{c}{PGD;eps=4/255} & \multicolumn{2}{c}{PGD;eps=2/255} \\
    \midrule
    \multicolumn{3}{c}{PGD-AT; eps=8/255} & \multicolumn{3}{c}{50.63} & \multicolumn{2}{c}{49.9} & \multicolumn{2}{c}{50.09} & \multicolumn{2}{c}{50.34} \\
    \multicolumn{3}{c}{PGD-AT; eps=4/255} & \multicolumn{3}{c}{50.87} & \multicolumn{2}{c}{50.87} & \multicolumn{2}{c}{50.87} & \multicolumn{2}{c}{50.87} \\
    \multicolumn{3}{c}{PGD-AT; eps=2/255} & \multicolumn{3}{c}{50.7} & \multicolumn{2}{c}{50.84} & \multicolumn{2}{c}{50.88} & \multicolumn{2}{c}{50.87} \\
    \multicolumn{3}{c}{PGD-AT; eps=1/255} & \multicolumn{3}{c}{50.8} & \multicolumn{2}{c}{50.85} & \multicolumn{2}{c}{50.87} & \multicolumn{2}{c}{50.88} \\
    \multicolumn{3}{c}{TRADES; eps=8/255} & \multicolumn{3}{c}{50.12} & \multicolumn{2}{c}{6.54} & \multicolumn{2}{c}{30.37} & \multicolumn{2}{c}{46.72} \\
    \multicolumn{3}{c}{TRADES; eps=4/255} & \multicolumn{3}{c}{49.12} & \multicolumn{2}{c}{18.01} & \multicolumn{2}{c}{35.64} & \multicolumn{2}{c}{46.58} \\
    \multicolumn{3}{c}{TRADES; eps=2/255} & \multicolumn{3}{c}{49.16} & \multicolumn{2}{c}{0.04} & \multicolumn{2}{c}{5.76} & \multicolumn{2}{c}{24.16} \\
    \bottomrule
    \end{tabular}
    }
  \label{tab:at}
\end{table}

\subsection{Detailed Experiment Settings}
In this section, we present the detailed settings of TRIM as described in Section 4.1.

TRIM consists of two main steps. The first step involves entropy prediction, which requires setting two threshold values $(\tau_{\text{low}}, \tau_{\text{high}})$ for the predicted entropy. The second step computes the KL divergence between the detector's outputs before and after random denoising. This step requires selecting an appropriate random denoising strategy $\mathcal{D}(\cdot)$ and setting a threshold $\tau_{\text{KL}}$ for KL divergence.

\paragraph{Detailed Settings for CNNSpot. }In our implementation of CNNSpot, we set the entropy thresholds to $\tau_{\text{low}} = 1 \times 10^{-15}$ and $\tau_{\text{high}} = 1 \times 10^{-1}$. For the random denoising strategy $\mathcal{D}(\cdot)$, we employ a combination of the following methods:
\begin{itemize}
    \item GaussianBlur with kernel size 3 and standard deviation $\sigma = 0.8$,
    \item RandomResizedCrop with scale range $(0.5,\ 1.0)$,
    \item RandomHorizontalFlip with $p=1$.
\end{itemize}
The threshold for KL divergence is set to $\tau_{\text{KL}} = 1$.

\paragraph{Detailed Settings for UnivFD. }In our implementation of UnivFD, we set the entropy thresholds to $\tau_{\text{low}} = 1 \times 10^{-6}$ and $\tau_{\text{high}} = 6 \times 10^{-1}$. For the random denoising strategy $\mathcal{D}(\cdot)$, we employ a combination of the following methods:
\begin{itemize}
    \item GaussianBlur with kernel size 3 and standard deviation $\sigma = 0.8$,
    \item RandomResizedCrop with scale range $(0.5,\ 1.0)$,
    \item RandomHorizontalFlip with $p=1$.
\end{itemize}
The threshold for KL divergence is set to $\tau_{\text{KL}} = 1$.

\paragraph{Detailed Settings for NPR. }In our implementation of NPR, we set the entropy thresholds to $\tau_{\text{low}} = 1 \times 10^{-25}$ and $\tau_{\text{high}} = 1 \times 10^{-1}$. For the random denoising strategy $\mathcal{D}(\cdot)$, Since blurring and resizing operations can disrupt the pixel-level correlations that NPR relies on, our random denoising adopts the following strategy:
\begin{itemize}
    \item RandomHorizontalFlip with $p=1$.
\end{itemize}
For the KL divergence threshold, we set it to $1 \times e^{-10}$ when the detector predicts the input as real, and $1 \times e^{-6}$ when the prediction is fake.

\paragraph{Detailed Settings for FreqNet. }In our implementation of FreqNet, we set the entropy thresholds to $\tau_{\text{low}} = 1 \times 10^{-20}$ and $\tau_{\text{high}} = 1 \times 5^{-1}$. For the random denoising strategy $\mathcal{D}(\cdot)$, Since Gaussian blurring and resizing operations can degrade the high-frequency information that FreqNet relies on, our random denoising adopts the following strategy:
\begin{itemize}
    \item RandomHorizontalFlip with $p=1$.
\end{itemize}
In FreqNet, we observe that the model's sensitivity to denoising varies depending on the dataset it is trained on. Therefore, we adopt dataset-specific KL divergence thresholds. For the FreqNet model trained on the ProGAN dataset, we set the KL threshold to $1 \times e^{-6}$ when the detector predicts the input as real, and $1 \times e^{-2}$ when the prediction is fake. For the model trained on the GenImage dataset, the threshold is set to $1 \times e^{-4}$ for real predictions and $1 \times e^{0}$ for fake predictions.

%%%%%%%%%%%%%%%%%%%%%%%%%%%%%%%%%%%%%%%%%%%%%%%%%%%%%%%%%%%%

%%%%%%%%%%%%%%%%%%%%%%%%%%%%%%%%%%%%%%%%%%%%%%%%%%%%%%%%%%%%
\bibliographystyle{unsrt} 
\bibliography{nips2025}

\begin{thebibliography}{10}

\bibitem{wang2020cnn}
Sheng-Yu Wang, Oliver Wang, Richard Zhang, Andrew Owens, and Alexei~A Efros.
\newblock Cnn-generated images are surprisingly easy to spot... for now.
\newblock In {\em Proceedings of the IEEE/CVF conference on computer vision and pattern recognition}, pages 8695--8704, 2020.

\bibitem{liu2020global}
Zhengzhe Liu, Xiaojuan Qi, and Philip~HS Torr.
\newblock Global texture enhancement for fake face detection in the wild.
\newblock In {\em Proceedings of the IEEE/CVF conference on computer vision and pattern recognition}, pages 8060--8069, 2020.

\bibitem{zhang2019detecting}
Xu~Zhang, Svebor Karaman, and Shih-Fu Chang.
\newblock Detecting and simulating artifacts in gan fake images.
\newblock In {\em 2019 IEEE international workshop on information forensics and security (WIFS)}, pages 1--6. IEEE, 2019.

\bibitem{frank2020leveraging}
Joel Frank, Thorsten Eisenhofer, Lea Sch{\"o}nherr, Asja Fischer, Dorothea Kolossa, and Thorsten Holz.
\newblock Leveraging frequency analysis for deep fake image recognition.
\newblock In {\em International conference on machine learning}, pages 3247--3258. PMLR, 2020.

\bibitem{zhang2023diffusion}
Yichi Zhang and Xiaogang Xu.
\newblock Diffusion noise feature: Accurate and fast generated image detection.
\newblock {\em arXiv preprint arXiv:2312.02625}, 2023.

\bibitem{ricker2024aeroblade}
Jonas Ricker, Denis Lukovnikov, and Asja Fischer.
\newblock Aeroblade: Training-free detection of latent diffusion images using autoencoder reconstruction error.
\newblock In {\em Proceedings of the IEEE/CVF Conference on Computer Vision and Pattern Recognition}, pages 9130--9140, 2024.

\bibitem{tan2023learning}
Chuangchuang Tan, Yao Zhao, Shikui Wei, Guanghua Gu, and Yunchao Wei.
\newblock Learning on gradients: Generalized artifacts representation for gan-generated images detection.
\newblock In {\em Proceedings of the IEEE/CVF Conference on Computer Vision and Pattern Recognition}, pages 12105--12114, 2023.

\bibitem{tan2024rethinking}
Chuangchuang Tan, Yao Zhao, Shikui Wei, Guanghua Gu, Ping Liu, and Yunchao Wei.
\newblock Rethinking the up-sampling operations in cnn-based generative network for generalizable deepfake detection.
\newblock In {\em Proceedings of the IEEE/CVF Conference on Computer Vision and Pattern Recognition}, pages 28130--28139, 2024.

\bibitem{liu2022detecting}
Bo~Liu, Fan Yang, Xiuli Bi, Bin Xiao, Weisheng Li, and Xinbo Gao.
\newblock Detecting generated images by real images.
\newblock In {\em European Conference on Computer Vision}, pages 95--110. Springer, 2022.

\bibitem{zhong2023patchcraft}
Nan Zhong, Yiran Xu, Sheng Li, Zhenxing Qian, and Xinpeng Zhang.
\newblock Patchcraft: Exploring texture patch for efficient ai-generated image detection.
\newblock {\em arXiv preprint arXiv:2311.12397}, 2023.

\bibitem{diao2024vulnerabilities}
Yunfeng Diao, Naixin Zhai, Changtao Miao, Zitong Yu, Xingxing Wei, Xun Yang, and Meng Wang.
\newblock Vulnerabilities in ai-generated image detection: The challenge of adversarial attacks.
\newblock {\em arXiv preprint arXiv:2407.20836}, 2024.

\bibitem{de2024exploring}
Vincenzo De~Rosa, Fabrizio Guillaro, Giovanni Poggi, Davide Cozzolino, and Luisa Verdoliva.
\newblock Exploring the adversarial robustness of clip for ai-generated image detection.
\newblock In {\em 2024 IEEE International Workshop on Information Forensics and Security (WIFS)}, pages 1--6. IEEE, 2024.

\bibitem{mavali2024fake}
Sina Mavali, Jonas Ricker, David Pape, Yash Sharma, Asja Fischer, and Lea Sch{\"o}nherr.
\newblock Fake it until you break it: On the adversarial robustness of ai-generated image detectors.
\newblock {\em arXiv preprint arXiv:2410.01574}, 2024.

\bibitem{saberi2023robustness}
Mehrdad Saberi, Vinu~Sankar Sadasivan, Keivan Rezaei, Aounon Kumar, Atoosa Chegini, Wenxiao Wang, and Soheil Feizi.
\newblock Robustness of ai-image detectors: Fundamental limits and practical attacks.
\newblock {\em arXiv preprint arXiv:2310.00076}, 2023.

\bibitem{dong2022think}
Chengdong Dong, Ajay Kumar, and Eryun Liu.
\newblock Think twice before detecting gan-generated fake images from their spectral domain imprints.
\newblock In {\em Proceedings of the IEEE/CVF conference on computer vision and pattern recognition}, pages 7865--7874, 2022.

\bibitem{hou2023evading}
Yang Hou, Qing Guo, Yihao Huang, Xiaofei Xie, Lei Ma, and Jianjun Zhao.
\newblock Evading deepfake detectors via adversarial statistical consistency.
\newblock In {\em Proceedings of the IEEE/CVF conference on computer vision and pattern recognition}, pages 12271--12280, 2023.

\bibitem{jia2022exploring}
Shuai Jia, Chao Ma, Taiping Yao, Bangjie Yin, Shouhong Ding, and Xiaokang Yang.
\newblock Exploring frequency adversarial attacks for face forgery detection.
\newblock In {\em Proceedings of the IEEE/CVF conference on computer vision and pattern recognition}, pages 4103--4112, 2022.

\bibitem{zhou2024stealthdiffusion}
Ziyin Zhou, Ke~Sun, Zhongxi Chen, Huafeng Kuang, Xiaoshuai Sun, and Rongrong Ji.
\newblock Stealthdiffusion: Towards evading diffusion forensic detection through diffusion model.
\newblock In {\em Proceedings of the 32nd ACM International Conference on Multimedia}, pages 3627--3636, 2024.

\bibitem{madry2019deeplearningmodelsresistant}
Aleksander Madry, Aleksandar Makelov, Ludwig Schmidt, Dimitris Tsipras, and Adrian Vladu.
\newblock Towards deep learning models resistant to adversarial attacks, 2019.

\bibitem{zhang2019theoreticallyprincipledtradeoffrobustness}
Hongyang Zhang, Yaodong Yu, Jiantao Jiao, Eric~P. Xing, Laurent~El Ghaoui, and Michael~I. Jordan.
\newblock Theoretically principled trade-off between robustness and accuracy, 2019.

\bibitem{tishby2000informationbottleneckmethod}
Naftali Tishby, Fernando~C. Pereira, and William Bialek.
\newblock The information bottleneck method, 2000.

\bibitem{robustbench}
Francesco Croce, Maksym Andriushchenko, Vikash Sehwag, Edoardo Debenedetti, Nicolas Flammarion, Mung Chiang, Prateek Mittal, and Matthias Hein.
\newblock Robustbench: a standardized adversarial robustness benchmark.
\newblock In Joaquin Vanschoren and Sai{-}Kit Yeung, editors, {\em NeurIPS Datasets and Benchmarks 2021}, 2021.

\bibitem{zhu2023genimage}
Mingjian Zhu, Hanting Chen, Qiangyu Yan, Xudong Huang, Guanyu Lin, Wei Li, Zhijun Tu, Hailin Hu, Jie Hu, and Yunhe Wang.
\newblock Genimage: A million-scale benchmark for detecting ai-generated image.
\newblock {\em Advances in Neural Information Processing Systems}, 36:77771--77782, 2023.

\bibitem{ojha2024universalfakeimagedetectors}
Utkarsh Ojha, Yuheng Li, and Yong~Jae Lee.
\newblock Towards universal fake image detectors that generalize across generative models, 2024.

\bibitem{achille2018emergence}
Alessandro Achille and Stefano Soatto.
\newblock Emergence of invariance and disentanglement in deep representations.
\newblock {\em Journal of Machine Learning Research}, 19(50):1--34, 2018.

\bibitem{achille2018information}
Alessandro Achille and Stefano Soatto.
\newblock Information dropout: Learning optimal representations through noisy computation.
\newblock {\em IEEE transactions on pattern analysis and machine intelligence}, 40(12):2897--2905, 2018.

\bibitem{amjad2019learning}
Rana~Ali Amjad and Bernhard~C Geiger.
\newblock Learning representations for neural network-based classification using the information bottleneck principle.
\newblock {\em IEEE transactions on pattern analysis and machine intelligence}, 42(9):2225--2239, 2019.

\bibitem{dogs_vs_cats}
{Microsoft}.
\newblock Dogs vs. cats.
\newblock \url{https://www.kaggle.com/c/dogs-vs-cats}, 2013.
\newblock Accessed: 2025-04-29.

\bibitem{croce2020reliableevaluationadversarialrobustness}
Francesco Croce and Matthias Hein.
\newblock Reliable evaluation of adversarial robustness with an ensemble of diverse parameter-free attacks, 2020.

\bibitem{pixle}
Jary Pomponi, Simone Scardapane, and Aurelio Uncini.
\newblock Pixle: a fast and effective black-box attack based on rearranging pixels.
\newblock In {\em 2022 International Joint Conference on Neural Networks (IJCNN)}, page 1–7. IEEE, July 2022.

\bibitem{andriushchenko2020squareattackqueryefficientblackbox}
Maksym Andriushchenko, Francesco Croce, Nicolas Flammarion, and Matthias Hein.
\newblock Square attack: a query-efficient black-box adversarial attack via random search, 2020.

\bibitem{tramer2020adaptiveattacksadversarialexample}
Florian Tramer, Nicholas Carlini, Wieland Brendel, and Aleksander Madry.
\newblock On adaptive attacks to adversarial example defenses, 2020.

\bibitem{carlini2017adversarialexampleseasilydetected}
Nicholas Carlini and David Wagner.
\newblock Adversarial examples are not easily detected: Bypassing ten detection methods, 2017.

\bibitem{rombach2022highresolutionimagesynthesislatent}
Robin Rombach, Andreas Blattmann, Dominik Lorenz, Patrick Esser, and Björn Ommer.
\newblock High-resolution image synthesis with latent diffusion models, 2022.

\bibitem{yu2015lsun}
Fisher Yu, Yinda Zhang, Shuran Song, Ari Seff, and Jianxiong Xiao.
\newblock Lsun: Construction of a large-scale image dataset using deep learning with humans in the loop.
\newblock In {\em Proceedings of the IEEE Conference on Computer Vision and Pattern Recognition (CVPR)}, 2015.

\bibitem{karras2018progressive}
Tero Karras, Timo Aila, Samuli Laine, and Jaakko Lehtinen.
\newblock Progressive growing of gans for improved quality, stability, and variation.
\newblock In {\em International Conference on Learning Representations (ICLR)}, 2018.

\bibitem{tan2024frequencyawaredeepfakedetectionimproving}
Chuangchuang Tan, Yao Zhao, Shikui Wei, Guanghua Gu, Ping Liu, and Yunchao Wei.
\newblock Frequency-aware deepfake detection: Improving generalizability through frequency space learning, 2024.

\bibitem{carlini2017evaluatingrobustnessneuralnetworks}
Nicholas Carlini and David Wagner.
\newblock Towards evaluating the robustness of neural networks, 2017.

\bibitem{croce2020minimallydistortedadversarialexamples}
Francesco Croce and Matthias Hein.
\newblock Minimally distorted adversarial examples with a fast adaptive boundary attack, 2020.

\bibitem{radford2021learningtransferablevisualmodels}
Alec Radford, Jong~Wook Kim, Chris Hallacy, Aditya Ramesh, Gabriel Goh, Sandhini Agarwal, Girish Sastry, Amanda Askell, Pamela Mishkin, Jack Clark, Gretchen Krueger, and Ilya Sutskever.
\newblock Learning transferable visual models from natural language supervision, 2021.

\bibitem{jin2023randomizedadversarialtrainingtaylor}
Gaojie Jin, Xinping Yi, Dengyu Wu, Ronghui Mu, and Xiaowei Huang.
\newblock Randomized adversarial training via taylor expansion, 2023.

\bibitem{Xu_2018}
Weilin Xu, David Evans, and Yanjun Qi.
\newblock Feature squeezing: Detecting adversarial examples in deep neural networks.
\newblock In {\em Proceedings 2018 Network and Distributed System Security Symposium}, NDSS 2018. Internet Society, 2018.

\bibitem{nie2022diffusionmodelsadversarialpurification}
Weili Nie, Brandon Guo, Yujia Huang, Chaowei Xiao, Arash Vahdat, and Anima Anandkumar.
\newblock Diffusion models for adversarial purification, 2022.

\bibitem{tang2024robust}
Linyu Tang and Lei Zhang.
\newblock Robust overfitting does matter: Test-time adversarial purification with fgsm.
\newblock In {\em Proceedings of the IEEE/CVF Conference on Computer Vision and Pattern Recognition}, pages 24347--24356, 2024.

\bibitem{schlarmann2024robustclipunsupervisedadversarial}
Christian Schlarmann, Naman~Deep Singh, Francesco Croce, and Matthias Hein.
\newblock Robust clip: Unsupervised adversarial fine-tuning of vision embeddings for robust large vision-language models, 2024.

\bibitem{liu2023forgeryawareadaptivetransformergeneralizable}
Huan Liu, Zichang Tan, Chuangchuang Tan, Yunchao Wei, Yao Zhao, and Jingdong Wang.
\newblock Forgery-aware adaptive transformer for generalizable synthetic image detection, 2023.

\bibitem{liu2024forgery}
Huan Liu, Zichang Tan, Chuangchuang Tan, Yunchao Wei, Jingdong Wang, and Yao Zhao.
\newblock Forgery-aware adaptive transformer for generalizable synthetic image detection.
\newblock In {\em Proceedings of the IEEE/CVF Conference on Computer Vision and Pattern Recognition}, pages 10770--10780, 2024.

\bibitem{moosavi2017universal}
Seyed-Mohsen Moosavi-Dezfooli, Alhussein Fawzi, Omar Fawzi, and Pascal Frossard.
\newblock Universal adversarial perturbations.
\newblock In {\em Proceedings of the IEEE conference on computer vision and pattern recognition}, pages 1765--1773, 2017.

\bibitem{abusnaina2021adversarial}
Ahmed Abusnaina, Yuhang Wu, Sunpreet Arora, Yizhen Wang, Fei Wang, Hao Yang, and David Mohaisen.
\newblock Adversarial example detection using latent neighborhood graph.
\newblock In {\em Proceedings of the IEEE/CVF international conference on computer vision}, pages 7687--7696, 2021.

\bibitem{ma2018characterizing}
Xingjun Ma, Bo~Li, Yisen Wang, Sarah~M Erfani, Sudanthi Wijewickrema, Grant Schoenebeck, Dawn Song, Michael~E Houle, and James Bailey.
\newblock Characterizing adversarial subspaces using local intrinsic dimensionality.
\newblock {\em arXiv preprint arXiv:1801.02613}, 2018.

\bibitem{zhang2024detecting}
Chi Zhang, Wenbo Zhou, Kui Zhang, Jie Zhang, Weiming Zhang, and Nenghai Yu.
\newblock Detecting adversarial examples via reconstruction-based semantic inconsistency.
\newblock In {\em Proceedings of the ACM Turing Award Celebration Conference-China 2024}, pages 126--131, 2024.

\bibitem{dolatabadi2024devil}
Hadi~M Dolatabadi, Sarah Erfani, and Christopher Leckie.
\newblock The devil’s advocate: Shattering the illusion of unexploitable data using diffusion models.
\newblock In {\em 2024 IEEE Conference on Secure and Trustworthy Machine Learning (SaTML)}, pages 358--386. IEEE, 2024.

\bibitem{may2023salient}
Brandon~B May, N~Joseph Tatro, Dylan Walker, Piyush Kumar, and Nathan Shnidman.
\newblock Salient conditional diffusion for defending against backdoor attacks.
\newblock {\em arXiv preprint arXiv:2301.13862}, 2023.

\bibitem{wang2022guided}
Jinyi Wang, Zhaoyang Lyu, Dahua Lin, Bo~Dai, and Hongfei Fu.
\newblock Guided diffusion model for adversarial purification.
\newblock {\em arXiv preprint arXiv:2205.14969}, 2022.

\bibitem{zhou2022improvingadversarialrobustnessmutual}
Dawei Zhou, Nannan Wang, Xinbo Gao, Bo~Han, Xiaoyu Wang, Yibing Zhan, and Tongliang Liu.
\newblock Improving adversarial robustness via mutual information estimation, 2022.

\end{thebibliography}

\end{document}